\documentclass{article}

\usepackage{PRIMEarxiv}
\usepackage[utf8]{inputenc} 
\usepackage[T1]{fontenc}    
\usepackage{hyperref}       
\usepackage{url}            
\usepackage{booktabs}       
\usepackage{amsfonts}       
\usepackage{nicefrac}       
\usepackage{microtype}      
\usepackage{lipsum}
\usepackage{fancyhdr}       
\usepackage{graphicx}       
\graphicspath{{media/}}     

\usepackage{tabularx}
\usepackage{lipsum}
\usepackage{graphicx}
\usepackage{subfigure}
\usepackage{enumitem}
\usepackage{wrapfig}
\usepackage{amssymb}
\usepackage{pifont}
\usepackage[ruled,vlined]{algorithm2e}
\usepackage{mathtools}

\DeclarePairedDelimiter\floor{\lfloor}{\rfloor}
\usepackage{amsmath}

\newcommand{\model}[1]{\textit{ViFiT}}
\newcommand{\modelf}[1]{\textit{Vi-Fi-Former-10F}}
\newcommand{\modelfff}[1]{\textit{Vi-Fi-Former-30F}}
\newcommand{\modelfffff}[1]{\textit{Vi-Fi-Former-50F}}
\newcommand{\modelfffffff}[1]{\textit{Vi-Fi-Former-70F}}
\newcommand{\modelfffffffff}[1]{\textit{Vi-Fi-Former-90F}}
\newcommand{\modelfffffffffff}[1]{\textit{Vi-Fi-Former-110F}}

\newcommand{\systwoplus}[1]{\textit{Vi-Fi-Former-2+}}
\newcommand{\systwoplusf}[1]{\textit{Vi-Fi-Former-10F-2+}}
\newcommand{\systwoplusfff}[1]{\textit{Vi-Fi-Former-30F-2+}}
\newcommand{\systwoplusfffff}[1]{\textit{Vi-Fi-Former-50F-2+}}
\newcommand{\systwoplusfffffff}[1]{\textit{Vi-Fi-Former-70F-2+}}
\newcommand{\systwoplusfffffffff}[1]{\textit{Vi-Fi-Former-90F-2+}}
\newcommand{\systwoplusfffffffffff}[1]{\textit{Vi-Fi-Former-110F-2+}}

\newcommand{\systwo}[1]{\textit{Vi-Fi-Former-2}}
\newcommand{\systwof}[1]{\textit{Vi-Fi-Former-10F-2}}
\newcommand{\systwofff}[1]{\textit{Vi-Fi-Former-30F-2}}
\newcommand{\systwofffff}[1]{\textit{Vi-Fi-Former-50F-2}}
\newcommand{\systwofffffff}[1]{\textit{Vi-Fi-Former-70F-2}}
\newcommand{\systwofffffffff}[1]{\textit{Vi-Fi-Former-90F-2}}
\newcommand{\systwofffffffffff}[1]{\textit{Vi-Fi-Former-110F-2}}

\newcommand{\nmodel}[1]{\textit{ViFiT}}
\newcommand{\nmodelf}[1]{\textit{ViFiT-10F}}
\newcommand{\nmodelfff}[1]{\textit{ViFiT-30F}}
\newcommand{\nmodelfffff}[1]{\textit{ViFiT-50F}}
\newcommand{\nmodelfffffff}[1]{\textit{ViFiT-70F}}
\newcommand{\nmodelfffffffff}[1]{\textit{ViFiT-90F}}
\newcommand{\nmodelfffffffffff}[1]{\textit{ViFiT-110F}}

\newcommand{\nsystwoplus}[1]{\textit{ViFiT-2+}}
\newcommand{\nsystwoplusf}[1]{\textit{ViFiT-10F-2+}}
\newcommand{\nsystwoplusfff}[1]{\textit{ViFiT-30F-2+}}
\newcommand{\nsystwoplusfffff}[1]{\textit{ViFiT-50F-2+}}
\newcommand{\nsystwoplusfffffff}[1]{\textit{ViFiT-70F-2+}}
\newcommand{\nsystwoplusfffffffff}[1]{\textit{ViFiT-90F-2+}}
\newcommand{\nsystwoplusfffffffffff}[1]{\textit{ViFiT-110F-2+}}

\newcommand{\nsystwo}[1]{\textit{ViFiT-2}}
\newcommand{\nsystwof}[1]{\textit{ViFiT-10F-2}}
\newcommand{\nsystwofff}[1]{\textit{ViFiT-30F-2}}
\newcommand{\nsystwofffff}[1]{\textit{ViFiT-50F-2}}
\newcommand{\nsystwofffffff}[1]{\textit{ViFiT-70F-2}}
\newcommand{\nsystwofffffffff}[1]{\textit{ViFiT-90F-2}}
\newcommand{\nsystwofffffffffff}[1]{\textit{ViFiT-110F-2}}

\newcommand{\baselinemodel}[1]{\textit{X-Translator}}
\newcommand{\baselinemodelf}[1]{\textit{X-Translator-10F}}
\newcommand{\baselinemodelfff}[1]{\textit{X-Translator-30F}}
\newcommand{\baselinemodelfffff}[1]{\textit{X-Translator-50F}}
\newcommand{\baselinemodelfffffff}[1]{\textit{X-Translator-70F}}
\newcommand{\baselinemodelfffffffff}[1]{\textit{X-Translator-90F}}
\newcommand{\baselinemodelfffffffffff}[1]{\textit{X-Translator-110F}}
\newcommand{\baselinesys}[2]{\textit{ViTag}}
\newcommand{\baselinesysf}[2]{\textit{ViTag-10F}}
\newcommand{\baselinesysfff}[2]{\textit{ViTag-30F}}
\newcommand{\baselinesysfffff}[2]{\textit{ViTag-50F}}
\newcommand{\baselinesysfffffff}[2]{\textit{ViTag-70F}}
\newcommand{\baselinesysfffffffff}[2]{\textit{ViTag-90F}}
\newcommand{\baselinesysfffffffffff}[2]{\textit{ViTag-110F}}
\newcommand{\dataset}[2]{Vi-Fi Dataset}

\newcommand{\cmark}{\ding{51}}%
\newcommand{\xmark}{\ding{55}}%

\usepackage{soul}

\newcommand{\bbc}[2] {\textcolor{cyan}{Bo:}}

\title{\textit{ViFiT}: Reconstructing Vision Trajectories from IMU and Wi-Fi Fine Time Measurements
}

\author{
  Author1, Author2 \\
  Affiliation \\
  Univ \\
  City\\
  \texttt{\{Author1, Author2\}email@email} \\
   \And
  Author3 \\
  Affiliation \\
  Univ \\
  City\\
  \texttt{email@email} \\
}

\author{
  Bryan Bo Cao \\
  Department of Computer Science \\
  Stony Brook University \\
  \texttt{boccao@cs.stonybrook.edu} \\
        \And
  Abrar Alali \\
  Old Dominion University \\
  Saudi Electronic University \\
  \texttt{a.alali@seu.edu.sa} \\
        \And
  Hansi Liu \\
  Rutgers University \\
  \texttt{hansiiii@winlab.rutgers.edu} \\
        \And
  Nicholas Meegan \\
  Rutgers University \\
  \texttt{njm146@scarletmail.rutgers.edu} \\
        \And
  Marco Gruteser \\
  Rutgers University \\
  \texttt{gruteser@winlab.rutgers.edu} \\
        \And
  Kristin Dana \\
  Rutgers University \\
  \texttt{kristin.dana@rutgers.edu} \\
        \And
  Ashwin Ashok \\
  Georgia State University \\
  \texttt{aashok@gsu.edu} \\
        \And
  Shubham Jain \\
  Stony Brook University \\
  \texttt{jain@cs.stonybrook.edu} \\
}

\begin{document}
\maketitle

\begin{abstract}
Tracking subjects in videos is one of the most widely used functions in camera-based IoT applications such as security surveillance, smart city traffic safety enhancement, vehicle to pedestrian communication and so on. 
In the computer vision domain, tracking is usually achieved by first detecting subjects with bounding boxes, then associating detected bounding boxes across video frames. For many IoT systems, images captured by cameras are usually sent over the network to be processed at a different site that has more powerful computing resources than edge devices. However, sending entire frames through the network causes significant bandwidth consumption that may exceed the system's bandwidth constraints. 
To tackle this problem, we propose \model~, a transformer-based model that reconstructs vision bounding box trajectories from phone data (IMU and Fine Time Measurements). It leverages a transformer's ability of better modeling long-term time series data. \model~ is evaluated on \dataset~~\footnote{Phone data is accessed only from opt-in participants who are part of this study. Data is used with the subjects' knowledge of opt-in consent.}, a large-scale multimodal dataset in 5 diverse real world scenes, including indoor and outdoor environments. To fill the gap of proper metrics of jointly capturing the system's characteristics of both tracking quality and video bandwidth reduction, we propose a novel evaluation framework dubbed Minimum Required Frames (MRF $\downarrow$) and Minimum Required Frames Ratio (MRFR $\downarrow$). \model~ achieves an MRFR of $0.65$ that outperforms the state-of-the-art approach for cross-modal reconstruction in LSTM Encoder-Decoder architecture \baselinemodel~ of $0.98$, resulting in a high frame reduction rate as $97.76\%$. 
\end{abstract}


\let\clearpage\relax
\section{Introduction}
\begin{figure*}[t]
  \begin{minipage}{1\linewidth}
  \centering
    \subfigure[Missing Frame]{\includegraphics[width=0.4285\textwidth]{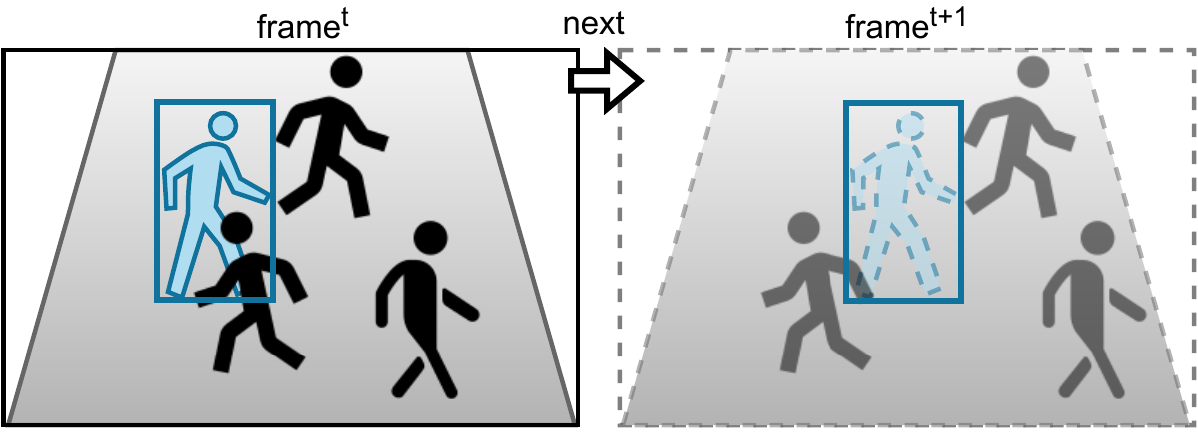}}
    \hspace{10pt}
    \subfigure[Missing Salient Part]{\includegraphics[width=0.4725\textwidth]{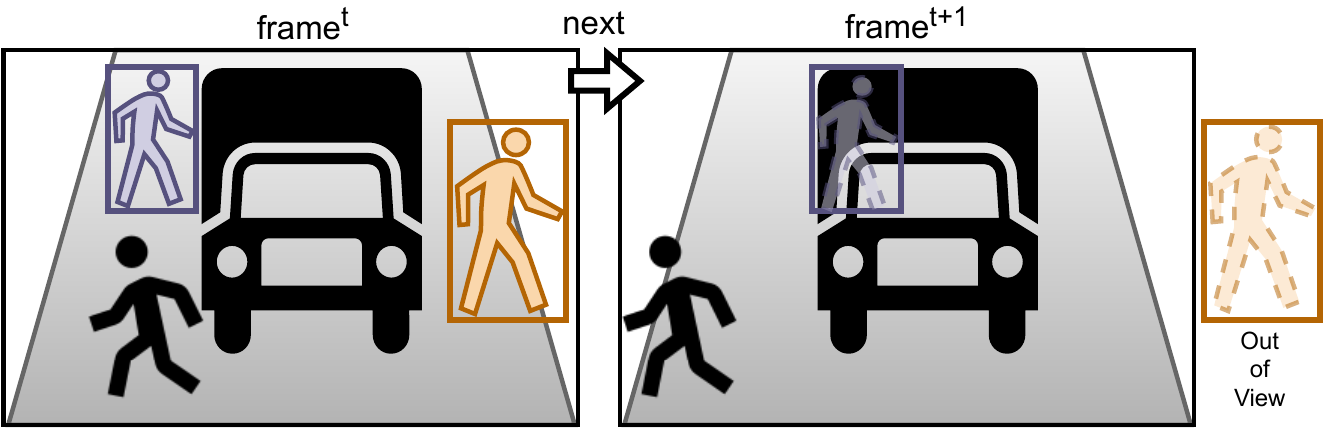}}
  \caption{Motivation. Two types of challenges using vision-only methods: \textbf{(a) Frame Drop}, an entire frame in the next timestamp is not available (e.g. due to temporal down sampling to save network bandwidth, network losses, etc.), resulting in missing visual information for estimating object of interests' detections (cyan); \textbf{(b) Salient Part Missing}: salient parts of objects are missing due to occlusion in the environment (purple) such as the truck or moving out of the camera's view (orange). Missing parts are displayed in lower opacity by dotted lines. Each color represents one identity of subject of interest. Detection ground truths are shown by solid bounding boxes.}
  \label{fig:datastudy}
  \end{minipage}
\end{figure*}

Tracking of human subjects in camera videos plays a key role in many real world Artificial Intelligence enabled Internet of Things (AIoT) applications, such as security surveillance, accident prevention, and traffic safety. State-of-the-art visual trackers rely on visual information from cameras but fail in scenarios with limited visibility, caused by poor light conditions, occlusion of the objects being tracked, or in out-of-view regions. 
Moreover, cameras installed for surveillance or other applications typically send all the image frames from the footage to a remote location for processing. This high volume of video data imposes constraints on network bandwidth. While it is possible to downsample the number of frames sent over the network, it is not desirable for applications such as tracking, that may require fine-grained information and can be negatively impacted by missing frames. This creates an inherent tradeoff between the requirement to preserve network bandwidth and the fidelity of the camera video.  Limiting the number of frames can preserve network bandwidth, however, will lead to missed image frames and thus tracking errors. There arises a clear need to reconstruct visual trajectories. 
To address this issue and the limitations of tracking based on vision only, prior works have leveraged other complementary modalities, such as raw sensory or meta data from phone and wireless signals, primarily through multimodal association~\cite{liu2022vi, cao2022vitag, meegan2022vificon}. However, if the camera image frames are lost or missing, then these approaches also fail in tracking.

To address this gap, we propose \model~, a system that can reconstruct a human subject's motion trajectory in camera video footage by leveraging motion sensor data from the subject's phone. Specifically, we capture inertial measurement unit (IMU) and Wi-Fi Fine Time Measurements (FTM) readings from the subjects' phones to reconstruct their vision trajectories. By leveraging lightweight modalities from the phone, we can ensure continuity in tracking information even when the camera frames are missing or the subjects are occluded or out of camera view. Our proposed approach also identifies the minimum number of camera frames required for vision trajectory reconstruction. In the future, \model~ can also serve as an adaptive downsampling technique to reduce the amount of camera data transmitted over the network. For example, as shown in Table~\ref{tab:datasize}, in a 10-frame window, video frames account for the majority of data (more than $99\%$) sent over the network. Therefore, by using \model~ we can significantly reduce the amount of information sent over the network.

\begin{table}[h]
  \begin{center}
    {\small{
\begin{tabular}{c|c|c@{\hskip1pt}|c@{\hskip6pt}c@{\hskip1pt}c}
\toprule
Domain & Modality & Data Size (kB) & Ratio \\
\midrule
Phone & IMU & 0.144 & $0.01\%$ \\
Phone & FTM & 0.144 & $0.01\%$ \\
\midrule
\textbf{Camera} & Vision (PNG) & 14,320 &  $\textbf{99.99}\% $\\
\textbf{Camera} & Vision (JPEG) & 1016 &  $\textbf{99.97}\% $\\
\bottomrule
\end{tabular}
}}
\end{center}
\caption{An empirical mesaurement of data transmitted through networks in a 10-frame window by sys.getsizeof() in Python. In the same time window, visual data from the camera domain dominates all data for over $99\%$ during network transmission.}
\label{tab:datasize}
\end{table}

\subsection{Challenges and Approach}

\begin{figure}[ht!]
\begin{center}
   \includegraphics[width=0.9\linewidth]{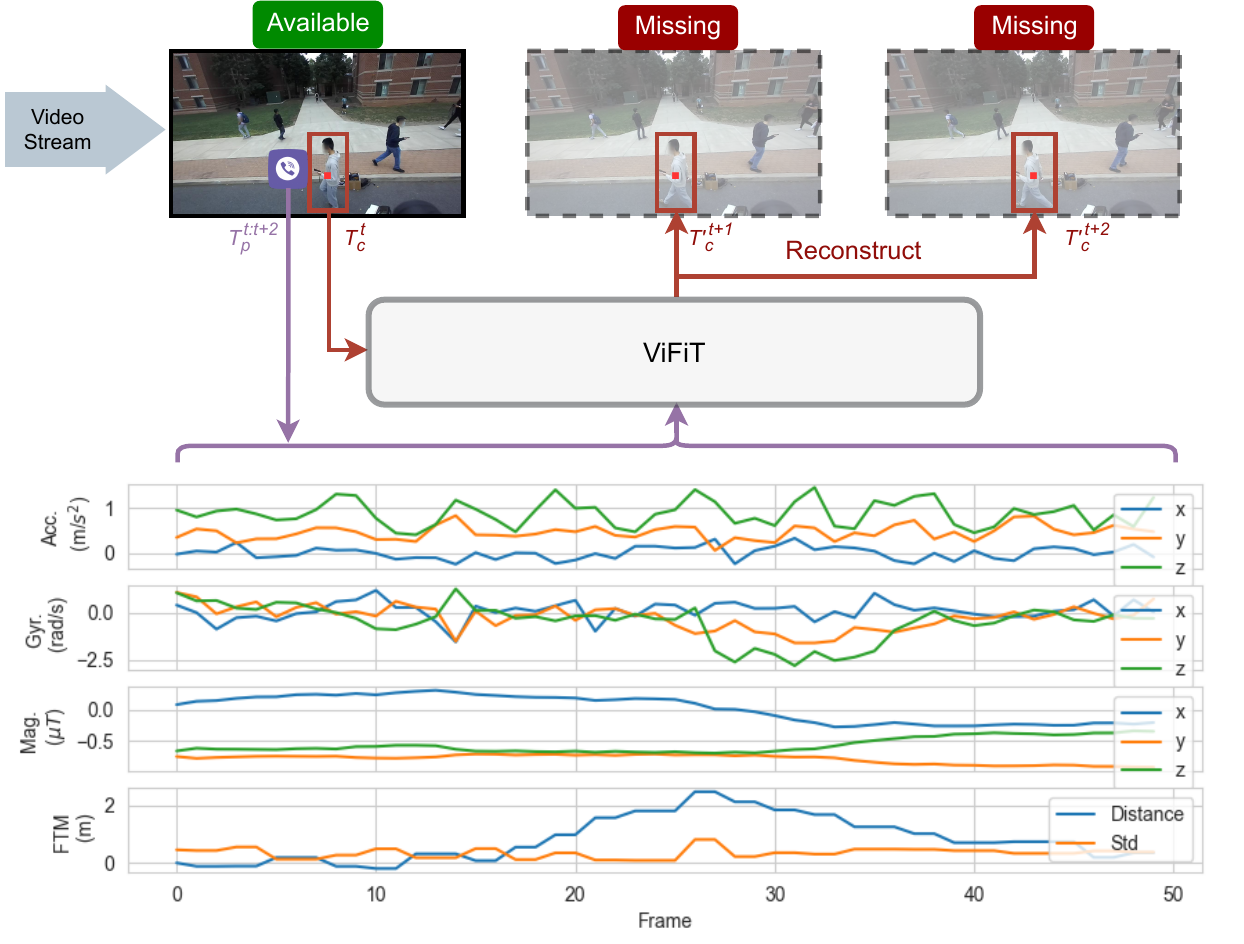}
\end{center}
   \caption{Task formulation: \nmodel~ reconstructs bounding boxes on top of missing frames by using complementary phone data including accelerations, gyroscope, magnetometer readings and wireless FTM data.}
\label{fig:mot}
\end{figure}

As illustrated in Fig.~\ref{fig:mot}, \model~ relies on an approach to reconstructing the tracklets (tracking coordinates on an image of the subject of interest) from the subject's phone's IMU and Wi-Fi data. This moves the onus of tracking to the phone's domain, yet also adds the challenge of associating the coordinates with the appropriate image frames and positioning to accurately mark the tracked subject. 
There are several challenges in reconstructing visual trajectories from phone data: (1) \textit{\textbf{Coordinate frame transformation}}: the phone and camera have different reference coordinate frames and therefore translating from one to the other requires a coordinate frame alignment. A naive extension of IONet ~\cite{chen2018ionet} that converts trajectories in a map representation to image coordinates will fail due to new challenges in generalizing to various cameras especially when their camera parameters are unknown; 
(2) \textit{\textbf{Multimodal fusion}}: the phone data includes raw IMU sensor readings and WiFi FTM values (distance from the access point), which have to be associated with camera image frames through a unified deep learning tracking model; (3) \textit{\textbf{IMU cumulative drift}}: also referred to as system errors in the literature, trajectories computed using IMU data are known to drift over time, potentially increasing errors in the reconstructed vision trajectories.
Recent development in phone data processing, especially IMU, has utilized deep learning models in various applications such as human activity recognition (HAR) ~\cite{jiang2015human, saeed2019multi, yao2017deepsense, xu2021limu}~, pedestrian multimodal association ~\cite{liu2022vi, cao2022vitag, meegan2022vificon} and etc. Sequential models have advanced from RNN, LSTM ~\cite{chen2018ionet, liu2022vi, cao2022vitag, meegan2022vificon} to the recently used transformer architecture ~\cite{xu2021limu}. Keeping in line with the state-of-the-art trackers using transformer models, in this paper we investigate, experiment and evaluate a transformer model with careful designs for vision trajectories reconstruction tasks in the images by using minimal image frames and multiple modalities of phone IMU and FTM data. 

The need to optimize the number of image frames under processing also has a connotation from a high-performance cluster computing resources perspective. It is becoming more common to use shared CPU and GPU resources across organizations and entities (or large-scale applications) to process videos. To speed up computation researchers have tried different approaches, particularly focusing on optimizing the necessary amount of video data. For example, video compression algorithm H.264, resolution reduction, or dropping frames. These approaches, however, reduce image quality or remove necessary information completely. As a result, tracking performance is degraded.
An alternate way is to reduce the video sampling rate, however, as the video stream is downsampled, we lose the subject's fine-grained movements across consecutive frames. Thus, the detected trajectories from a downsampled stream will likely be error-prone. 

We conducted a study to investigate the amount of data reduced (denoted as Data $\downarrow$ (MB)) as a result of skipping frames. When multiple cameras share a common GPU resource, reducing the amount of processing data for each camera can potentially allow the GPU resource to be shared across a larger number of cameras.
We preserve the first frame and drop the following frames in a window. Skipping frames saves a large amount of video data, i.e. in a 30 frames window, we can reduce 41.53MB data in PNG format, shown in Table~\ref{tab:datasizewin}.
Since phone data accounts for extremely small portion in a window ($0.02\%$), this video stream scaling factor is approximately the same as the ratio of skipped frames in a window. That is, when we skip 29 frames in a 30-frame window, the saving can facilitate the use of additional 29 video streams without scaling GPU resources. These preliminary explorations and compute optimization necessities further attest our proposed need to minimally use image frames, and leverage other modalities, to achieve the target fidelity.

\begin{table}[ht!]
  \begin{center}
    {\small{
\begin{tabular}{c|ccc@{\hskip2pt}c@{\hskip2pt}}
\toprule
\textit{{F}/{Win}} & Video Data (MB) & Format & Data $\downarrow$ (MB) & Cam Scale  \\
\midrule
10 & $1.02$ & JPEG & $-0.91$ & 10x\\
30 & $3.05$ & JPEG & $-2.95$ & 30x\\
50 & $5.08$ & JPEG & $-4.98$ & 50x\\
\midrule
10 & $14.32$ & PNG & $-12.89$ & 10x\\
30 & $42.96$ & PNG & $-41.53$ & 30x\\
50 & $71.60$ & PNG & $-70.17$ & 50x\\
\midrule
\textit{F} & Size$_{frame} \times$ \textit{F} & - & $-$Size$_{frame} \times$ (\textit{F} - 1) & \textit{F}x\\
\bottomrule
\end{tabular}
}}

\end{center}
\caption{Effect of frame dropping on network saving and camera streams scaling. $F/Win$: $\#$ frames in a window. Size of JPEG$_{frame}$=101.6KB=0.1016MB. Data $\downarrow$ indicates the amount of data to drop when we only send one frame in a window, resulting in the scaling of \#Cameras by a factor of $F$x.}
\label{tab:datasizewin}
\end{table}

\noindent \textbf{Contributions.} In summary, the contributions from this paper are as follows:
\begin{itemize}[leftmargin=*,noitemsep,topsep=0pt]
    \item 
    We design and develop \model~, a novel multimodal transformer-based model to reconstruct vision tracklets from phone domain data, including IMU and FTM data.
    \item
    We develop a novel Intersection over Union (IoU) based metric called {\em Minimum Required Frames} (MRF) that jointly captures both the quality of reconstructed bounding boxes and the lower bound of frames required for reconstruction by the complementary phone data in video streams.
    \item
    \nmodel~ achieves an MRF of 33.75 in all 4 outdoor scenes on average, outperforming the best baseline method with an MRF of 53 ($\Delta=$ 15.25). \model~ uses an extremely small amount of video frames with only $2.24\%$, demonstrating its effectiveness to reconstruct accurate bounding boxes (IoU>0.5) with few frames needed in a video stream (described in details in Fig. \ref{fig:MRF}).
\end{itemize}

\section{Related Work}
\noindent \textbf{Vision-based Detection \& Tracking} There is a large body of research on object detection ~\cite{girshick2015fast, redmon2016you, redmon2017yolo9000, redmon2018yolov3, bochkovskiy2020yolov4, glennjocher} and tracking ~\cite{fan2018deep, gao2020multiple}. Recent years has seen an emerge of research on transformer-based models in a wide variety of visual tasks, such as image classification ~\cite{dosovitskiy2020image}, object detection ~\cite{carion2020end}, tracking ~\cite{chen2021transformer} and so forth. 
Common vision benchmarks in include COCO ~\cite{lin2014microsoft} for object detection, MOT ~\cite{milan2016mot16} for tracking. However, these are for vision-only evaluation when all the frames in a complete video are available. Li et al. investigate the evaluation in case of frame dropping ~\cite{li2020towards}. In another line of research, visual trajectory reconstruction datasets include BIWI ~\cite{pellegrini2009you}, UCY ~\cite{lerner2007crowds} and ~\cite{hug2021quantifying} but lack phone modality. We hereby use the Vi-Fi dataset ~\cite{vifisite} ~\cite{liu2022vi} that include vision, IMU and FTM for our research.

Though both objector detectors and our work yield the same output -- bounding boxes in image coordinates, there are several challenges in order to make fair comparisons. Key differences are summarized as follows: (1) salient objects can be located precisely in pixel locations of an image; in contrast, we regress objects' bounding boxes in future frames of a window relative to the first frame from IMU data, without objects location information in future image coordinates. (2) missing future frames do not have any visual information, thus pure vision object detectors trackers will fail to produce any detections, while our model can continuously produce bounding boxes. (3) CV object detectors detect objects on each frame independently, while positions estimated from IMU readings have notorious drifts cumulated in a series of data \cite{chen2018ionet}, making bounding boxes deviate from ground truths in a longer time sequence. The latter inevitably deteriorates the performance of future bounding box regression in terms of  Average Precision (AP) and IoU scores.

\noindent \textbf{Multimodal Learning and Fusion.} Deep learning has facilitated a wide range of applications using IMU data, including human activity recognition ~\cite{reyes2016transition, jiang2015human, saeed2019multi}, trajectory estimation from IONet ~\cite{chen2018ionet}, phone domain transformation ~\cite{chen2019motiontransformer}, etc. Another line of research has investigated the fusion of multiple modalities of vision, including,
human identification by wifi and vision ~\cite{deng2022gaitfi}, localization and tracking by vision, IMU or network ~\cite{zhou2012long, jiang2018ptrack, zhao2020urban, dong2021enabling}, egomotion estimation ~\cite{lu2020milliego}. Prior to learning-based method, Kalman filter ~\cite{welch1995introduction, li2015kalman} is commonly used for tracking.

The closest works to our research from multimodal learning aspect include IONet ~\cite{chen2018ionet} and RoNIN ~\cite{herath2020ronin} for regressing trajectories on a map representation from IMU, Vi-Fi ~\cite{liu2022vi}, ViTag ~\cite{cao2022vitag} and ViFiCon ~\cite{meegan2022vificon} for associating camera and phone data, as well as PK-CEN ~\cite{chen2020pedestrian} and FK-CEN ~\cite{xiao2022pedestrian} for visual tracking. However, there are several key differences. The first three focus on multimodal association. For the last two, the authors apply LSTM-based models with neighboring heterogeneous traffic information to predict future frames, while they exploit pose and even facial keypoints for future trajectory prediction, respectively. The latter poses serious privacy concerns due to the use of biometric data of facial information. By using pose or human mesh representation ~\cite{das2020vpn}, pose-based video surveillance can alleviate the privacy problem. Zheng et al. propose to learn a pose-invariant embedding for person-reidentification ~\cite{zheng2019pose}. We refer readers to the details in the pose estimation survey ~\cite{zheng2020deep}. Our system differs in that we only utilize motion information that treats each pedestrian as a moving point without any facial or pose information, which hinders the leakage of user's identity information to preserve privacy. In addition, in PK-CEN ~\cite{chen2020pedestrian} and FK-CEN ~\cite{xiao2022pedestrian}, a historical 8 samples from vision (3.2s) has to be known while only one frame is available in our work. Moreover, \model~ works in a more challenging scenario that the coordinate transformation between camera and real world is not explicitly given. We summarize the main characteristics in Table ~\ref{tab:trajimu} and ~\ref{tab:pk}.

\begin{table}[h]
  \begin{center}
    {\small{
\begin{tabular}{c|cccc}
\toprule
Method & Window Length (WL) & Input & Output \\
\midrule
IONet ~\cite{chen2018ionet} & 2s200f & (a, g)$_{200\times6}$ & ($\Delta l, \Delta \psi$)$_{1\times2}$ \\
6-DOF ~\cite{silva2019end} & 2s200f & (a, g)$_{200\times6}$ & ($\Delta l, \Delta q$)$_{1\times7}$ \\
\textbf{\modelfff~} & 3s30f & ($T_{i}$, $T_{f}$)$_{30\times9}$ +(1f)$_{1\times5}$  &  ($T_{c}$)$_{30\times5}$ \\
\bottomrule
\end{tabular}
}}
\end{center}
\caption{Summary of IMU-based Trajectory Reconstruction in the Literature. Abbreviations: a - linear acceleration; g - angular velocity; l - position; $\psi$ - heading; q - quaternion; f - frame; s - second. In constrast to IONet and 6-DOF odometry that construct one positional datapoint in a window, \model~ performs in a more challenging task that it reconstructs a series of bounding boxes in all missing frames in a window. This task requires the knowledge of World-Camera coordinates transformation while it's not explicitly given in the Vi-Fi dataset ~\cite{liu2022vi}.}
\label{tab:trajimu}
\end{table}

\begin{table*}[h]
  \begin{center}
    {\small{
\begin{tabular}{c|ccccc}
\toprule
System & Frame Rate & Observed & Reconstructed & Input & Privacy \\
 & (FPS) $\uparrow$ & Frames $\downarrow$ & Frames $\uparrow$ & & \\
\midrule
PK-CEN ~\cite{chen2020pedestrian} & 2.5 & 8 & 12  & Pose kpt, social inter. \& near obj. & \xmark \\ 
FK-CEN ~\cite{xiao2022pedestrian} & 2.5 & 8 & 12 & Facial kpt, social inter. \& near obj. & \xmark\\

\textbf{\model~} & \textbf{10} & \textbf{1} & \textbf{29} &  Obj. cam coord., IMU \& FTM          & \cmark \\
\bottomrule
\end{tabular}
}}
\end{center}
\caption{Summary of system characteristics in recent works.} 
\label{tab:pk}
\end{table*}

Transformer has been used successfully in a wide range of modalities, including NLP ~\cite{vaswani2017attention}, vision ~\cite{dosovitskiy2020image}, IMU ~\cite{xu2021limu} and so forth. MissFormer ~\cite{becker2021missformer} also investigated reconstruction. However, it only focuses on self-reconstruction on unimodality - 2D trajectory. In contrast, our work take a step further to not only introduce modalities in the phone domain with IMU and FTM, but also reconstruct bounding boxes that in the camera domain. In summary, our work extends it to an novel multimodal fusion task on vision, IMU and FTM.

\section{Methodology}

\subsection{Data Preprocessing}
We utilize the large scale multimodal dataset collected in Vi-Fi ~\cite{liu2022vi}. Readers can refer to the website ~\cite{vifisite} for the details of data collection and format. Vi-Fi dataset consists of RGB-D (depth) visual data captured by a ZED-2 stereo camera, wireless data by communication between Google Pixel 3a phones and a Google Nest Wi-Fi access point next to the camera, as well as IMU accelerometer, gyroscope, magnetometer sensor readings from the phone. It covers a wide range of scenes both indoors and outdoors, totaling 142K frames in 89 sequences, each of which lasts around 3 minutes. At most 5 subjects are holding phones communicating with the access point and 11 detections in one scene. All participants walk in an unconstrained fashion.

\noindent \textbf{Camera Data.}
We follow the procedure in ViTag ~\cite{cao2022vitag} to generate trajectories (referred to as \textit{tracklets} in the rest of the paper) using the StereoLabs ZED tracker on the RGB-Depth camera data. Tracklets are typically short because subjects move out of the field of view of the camera frequently.
Tracklets from camera data ($T_c$) are represented as a time series sequence of bounding boxes (\textit{BBX}). Each bounding box is represented as:
\begin{equation}
BBX = [x, y, d, w, h], \quad T_{c} \in \mathbb{R}^{WL \times 5}
\label{eqn:bbx}
\end{equation}
\noindent where $x$ and $y$ are the coordinates of the centroid of the bounding box, $d$ is the centroid's depth measurement, and $w$ and $h$ are the bounding box width and height respectively. $WL$ is the window's length which is also the number of frames in a window.

\noindent \textbf{Phone Data.} To preprocess the smartphone data, we use 3 types of measurements from the time series IMU data. We extract the 3-axis accelerometer data $acc$, 3-axis gyroscope $gyro$ and magnetometer data $mag$. These measurements from the time series IMU data are concatenated as a vector:

\begin{equation}
    T_i^t = [acc; gyro; mag],
    \quad T_{i} \in \mathbb{R}^{WL \times 9}
    \label{eqn:imu}
\end{equation}
\noindent
Additionally, we use phones' FTM measurements at time $t$ which is defined as:
\begin{equation}
    T_f^t = [r, std],
    \quad T_{f} \in \mathbb{R}^{WL \times 2}
    \label{eqn:ftm}
\end{equation}
where $r$ represents the estimated range, or distance from phone to WiFi access point, while $std$ represents the standard deviation calculated in a single RTT burst.

In the context of our work, we use \textit{modality} to refer to one type of data such as bounding boxes, IMU readings, or FTM data, while we use \textit{domain} to refer to the source, such as camera or smartphone. Thus, vision tracklets ($T_c$) belong to the camera domain, and IMU and FTM data belong to the phone domain ($T_{p}$):
\begin{equation}
    T_{p} = [T_{i}; T_{f}]
    \label{eqn:p}
\end{equation}

Different from ViTag ~\cite{cao2022vitag} when the task is performed with a fixed window length ($WL$) that are always less than 3 seconds, we provide a \model~ family with various $WL$s that can meet a variety of practical duration constraints.

\noindent \textbf{Multimodal Synchronization.} 
Due to the variety of sampling rates and timestamps, we synchronize all the modalities before feeding them to the model. We use Network Time Protocol (NTP) on the devices to synchronize the camera and phone data. The sampling rate for camera frames is 30 FPS, for IMU readings is 100 Hz, and 3-5 Hz for FTM. In addition, camera (BBX) and phone (IMU, FTM) data have $16$ and $13$ precision timestamps. To unify the sampling rates, we first downsample the data from the camera domain to 10 FPS. Then we synchronize the phone modalities using the camera timestamp as a reference. Specifically, we find the reading of IMU and FTM with the nearest timestamps to the camera frame timestamp.

\noindent \textbf{Normalization.} Real world IMU readings normally are captured with noise. Inspired by LIMU-BERT \cite{xu2021limu}, by default we apply normalization on IMU data by:
\begin{equation}
    acc_{{j}} = \frac{acc_{{j}}}{9.8}, \hspace{10pt} mag_{{j}}= \frac{\alpha mag_{{j}}}{\sqrt{\sum mag^{2}_{{j}}}}, \hspace{10pt} {j} \in \mathbb{R}^{3}
    \label{eqn:norm}
\end{equation}
where $acc_{{j}}$ and $mag_{{j}}$ are accelerometer and magnetometer sensor raw readings in the ${j}th$ axis. This way both acceleration and magnetometer data are distributed in similar ranges. Since gyroscope readings are generally small, we do not perform any normalization. In another set of experiment, we apply Savitzky-Golay smoothing filter ~\cite{savitzky1964smoothing} with a window length of 11 and a degree 2 polynomial on each axis of all IMU features, including accelerameter, gyroscope and magnetometer data. Results are discussed in Table ~\ref{tab:norm}.

\subsection{Inertial Trajectory Reconstruction}

We present the background of trajectory reconstruction from inertial data in this section.

The central problem is formulated as estimating the locations $L^{t}$ of a trajectory during a continuous period of time $t \in T$ given the inertial information of acceleration $acc$ and angular velocity $gyro$ from IMU sensors. As illustrated in IONet ~\cite{chen2018ionet}, this process is formulated as a function $f(\cdot)$ in Equation ~\ref{equ:imu2l}:

\begin{equation}
    [\textbf{O}_{b \rightarrow n} \quad \textbf{v} \quad \textbf{L}]^{t}  = f([\textbf{O}_{b \rightarrow n} \quad \textbf{v} \quad \textbf{L}]^{t-1}, [\textbf{acc}, \textbf{gyro}]^{t})
    \label{equ:imu2l}
\end{equation}

where $O_{b \rightarrow n}$ is the direction cosine matrix from body frame ($b$) to navigation ($n$) coordinates, $v$ is velocity. Note the states of attitude, velocity and location $[\textbf{O}_{b \rightarrow n} \ \ \textbf{v} \ \ \textbf{L}]^{t}$  at time $t$ cannot be directly observed, which requires the derivation from both previous system states $[\textbf{O}_{b \rightarrow n} \ \textbf{v} \ \textbf{L}]^{t-1}$ at time $(t-1)$ and measurments from inertial sensors $[\textbf{acc}, \textbf{gyro}]^{t}$ at time $t$. In traditional Strapdown Inertial Navigation System (SINS) ~\cite{savage1998strapdown}, by basic Newtonian Laws of Motion we can infer that acceleration on the horizontal plane $\bar{acc}$ can be derived by subtracting $acc$ from the gravity vector $g$. Then velocity is calculated by integration of $\bar{acc}$, while distance (the cumulative change of locations) is estimated by double integration of $\bar{acc}$.

Now consider an independent window of $WL$ frames, the displacement in L2 norm is express as:

\begin{equation}
    \Delta l = || \textbf{L} || = ||WL \textbf{v}_{b}^{0} \textit{dt} + \textbf{T} \textit{dt}^{2} - \frac{WL(WL - 1)}{2}\textbf{g}^{0}_{b}\textit{dt}^{2} ||_{2}
\end{equation}
where
\begin{equation}
    \textbf{T} = (WL - 1)acc^{0} + (WL-2)\Omega^{1}acc^{1} + ... + \prod_{w=1}^{WL-2}\Omega^{w}acc^{WL - 1}
\end{equation}
and $\textbf{v}_{b}^{0}$, $\textbf{g}^{0}_{b}$ $\Omega^{t}$ are the initial velocity, gravity at the first frame (0-indexed) and relative rotation matrix, respectively. Then a polar vector $(\Delta l, \Delta \psi)^{t}$ that represents the displacement and change of orientation can be described as function $f'(\cdot)$:
\begin{equation}
    (\Delta l, \Delta \psi)^{t} = f'(\textbf{v}_{b}^{0}, \textbf{g}_{b}^{0}, \bar{\textbf{acc}}^{2:WL-1}, \bar{\textbf{gyro}}^{2:WL-1})
\end{equation}
Next, the horizontal position at any time can be derived given the initial location:

\begin{equation}
          x^{t} = x^{0} + \Delta l cos(\psi^{0} + \Delta \psi) \\
    \label{equ:x_t}
\end{equation}
\begin{equation}
          y^t = y^{0} + \Delta l sin(\psi^{0} + \Delta \psi) \\
    \label{equ:y_t}
\end{equation}
Finally, the trajectory is simply a vector of locations $(x, y)$ from $t=0$ to $t=WL-1$.

Though in theory, a trajectory reconstruction system can be implemented only by function $f'(\cdot)$, the noise and errors are empirically too large when it comes to real world data captured from commercial sensors. IONet \cite{chen2018ionet} demonstrates that the function $f'(\cdot)$ can be learned by an RNN-based deep learning model that outperforms hand-crafted methods. We show in later section that transformer can even better learn the dynamics due to its ability to learn long-range dependency, which can help reduce cumulative drift from IMU sensors.

\subsection{ViFiT Design}
\begin{figure*}[t]
\begin{center}
   \includegraphics[width=\linewidth]{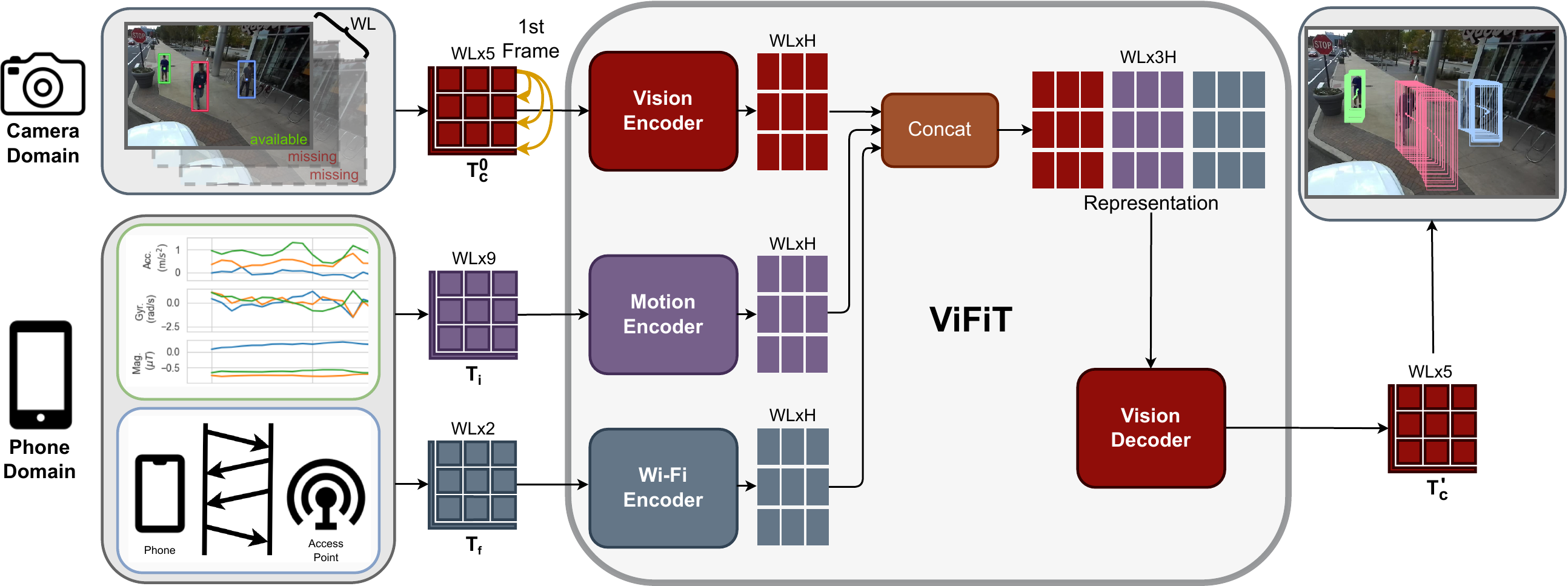}
\end{center}
   \caption{\model~ System Overview. \model~ consists of multimodal Encoders for ($T_{c}^{0}$, $T_{i}$ and $T_{f}$) to extract features and Vision Decoder to reconstruct the whole visual trajectory of $T_{c}'$ for the missing frames in a window with length $WL$. Note $T_{c}^{0}$ denotes a vision tracklet with first frame only and $H$ denotes representation dimension.}
\label{fig:datasize_dist}
\end{figure*}

\begin{figure*}[t]
\begin{center}
   \includegraphics[width=\linewidth]{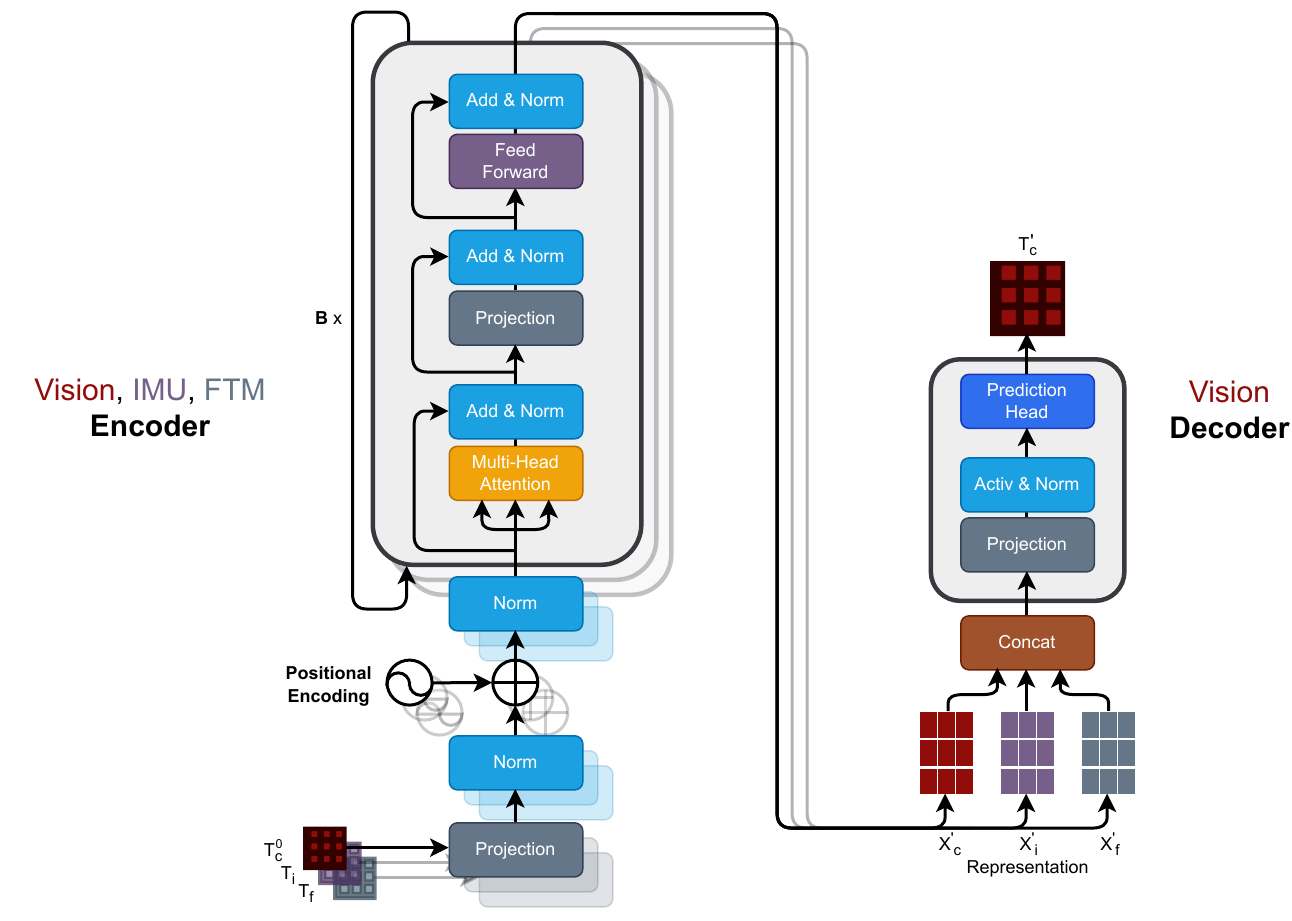}
\end{center}
   \caption{Vi-Fi Transformer (\model~) Architecture. \model~ is comprised of multimodal Encoders for ($T_{c}^{0}$, $T_{i}$ and $T_{f}$) depicted on the left side in parallel displayed with various degrees of opacity, as well as a Vision Decoder on the right. Information flow starts from the bottom left corner, where each tracklet for one modality ($T_{c}^{0}$, $T_{i}$ or $T_{f}$) is fed into its own Encoder independently, including $B$ blocks of transformer modules with Multi-head Self-attention (MSA). In the next step, Encoders generate multimodal representations, fused by concatenation ($X'_{c}$, $X'_{i}$, $X'_{f}$) and are fed into the Vision Decoder to output bounding boxes ($T_{c}^{\prime}$) in missing frames.}
\label{fig:vifit}
\end{figure*}

In this section, we describe the details of Vi-Fi Transformer\footnote{The term ``Vi-Fi Transformer'' is also referred to as ``Vi-Fi-Former'' in this paper.} -- \nmodel~. The workflow is illustrated in Fig. \ref{fig:vifit}~. We employ the main backbone of the transformer architecture inspired by ~\cite{xu2021limu} and ~\cite{vaswani2017attention} with a few key modifications: (1) we implement three separate independent encoders to learn modality-specific features; (2) intermediate representations are concatenated to fuse multiple modality information; and (3) residual connections between encoder and decoder are removed such that decoder purely depends on the fused multimodal representations.

\nmodel~ consists of three encoders for multiple modalities and one vision decoder. The objective of the encoders is to learn multimodal representations from vision, IMU and wireless data FTM, e.g. IMU Encoder is responsible for capturing motion information from accelerations, rotations, and orientations while FTM Encoder focuses on wireless data. We keep encoders identical for all modalities for the simplicity to extend new modalities such as RF in the future work. In the next step, representations are extracted and fused by concatenation, based on which the Vision Decoder reconstructs the bounding boxes.

\noindent \textbf{Encoder.} Each encoder comprises $B=4$ stacks of Multi-head Self-attention (MSA), Projection and Feed Forward layers, where residual connections and Layer Normalization are employed in between. It takes in a tracklet $T_{m}$ as input and projects it to a higher dimensional space using:

\begin{equation}
    X = Proj(T_{m}) = A \times T_{m}    
    \label{eqn:proj}
\end{equation}

\noindent where $A$ is a matrix with dimension $H_{dim} \times D_{m}$ and $H_{dim}$ is the hidden space dimension larger than modality $m$'s feature dimension $D_{m}$ (e.g. $D_{c}=5$, $D_{i}=9$ and $D_{f}=2$). We set $H_{dim}=72$ by default as it yields the best overall performance shown in later experimental sections. We implement the Projection layer \textit{Proj($\cdot$)} by a linear layer. The objective is to expand the low dimensional input feature space $D_{m}$ to a larger one to learn richer implicit features $X_{m}$ for modality $m$. Since data in different scenes have different distributions, it can lead to unstable training. Therefore, Layer Normalization ~\cite{ba2016layer} is applied to stabilize features of instance $i$ from previous Projection Layer:

\begin{equation}
    \hat{X}_{m}^{i} = LayerNorm(X_{m}^{i}) = \frac{X_{m,j}^{i} - \mu_{j}}{\sqrt{\sigma_{j}^{2} + \epsilon}}\gamma + \beta    
    \label{eqn:ln}
\end{equation}

\noindent where $\gamma$ and $\beta$ are the learnable hyperparameters, and $\epsilon$ is a small number to avoid numerical instability. The mean and standard deviation across modality $m$'s feature $j$ ($jth$ column of $A$) are denoted by $\mu_{j}$ and $\sigma_{j}$, respectively. After that, different from recurrent layers of LSTM in the Vi-Fi ~\cite{liu2022vi} or ViTag ~\cite{cao2022vitag} models, we employ positional encoding \cite{vaswani2017attention}~ learn the order information of modality $m$ and add it into $\hat{X}_{m}$, followed by a second Layer Normalization.

In the next step, $\hat{X}_{m}$ enters $B$ core transformer blocks with Multi-head Self-attention (MSA) layers. Scaled Dot-product Attention \cite{vaswani2017attention} is used with $d_{k}$ dimensional queries and keys while values are of dimension $d_{v}$, implemented by:
\begin{equation}
    MSA(\hat{X}_{m}) = MSA(\hat{Q}_{m}, \hat{K}_{m}, \hat{V}_{m}) = Concat(head_{1}, ..., head_{h})A_{m}^{O}
    \label{eqn:msa}
\end{equation}
\noindent where $A^{O} \in \mathbb{R}^{hd_{v} \times d_{model}}$ and a $head$ is an attention layer:
\begin{equation}
    Attention(\hat{X}_{m}) = Attention(\hat{Q}_{m}, \hat{K}_{m}, \hat{V}_{m}) = softmax(\frac{\hat{Q}_{m}\hat{K}_{m}^{T}}{\sqrt{d_{k}}})\hat{V}_{m}
    \label{eqn:atten}
\end{equation}
\noindent where $\hat{Q}_{m}=\hat{K}_{m}=\hat{V}_{m}=\hat{X}_{m}$ such that a modality's hidden features attend to its own values to learn the relative importance. More numbers of heads ($h$) allow for learning different representations. In this implementation we set $h=4$, $d_{k} = d_{v} = H_{dim} = 72$ and $d_{model}= H_{dim} \times h = 72 \times 4 = 288$.

The Position-wise Feed Forward layer (denoted as \textit{FFN($\cdot$)} and Feed Forward in Fig. \ref{fig:vifit}) is implemented by two linear transformations of dimension $H_{dim}=72$ and $F_{dim}=144$, respectively. Following ~\cite{xu2021limu} but different from \cite{vaswani2017attention}, Gaussian Error Linear Unit (GELU) ~\cite{hendrycks2016gaussian} is utilized as the activation function between two layers.

Putting all together, a transformer block is implemented by the following functions:
\begin{equation}
    M^{b} = LayerNorm(MSA(\hat{X}_{m}^{b - 1}) + \hat{X}_{m}^{b - 1})
    \label{eqn:tfmblock0}
\end{equation}
\begin{equation}
    P^{b} = LayerNorm(Proj(M^{b}) + M^{b})
    \label{eqn:tfmblock1}
\end{equation}
\begin{equation}
    \hat{X}_{m}^{b} = LayerNorm(FFN(P^{b}) + P^{b})
    \label{eqn:tfmblock}
\end{equation}
\noindent where $b - 1$ denotes a previous block. The final representation of modality $m$ from encoder is denoted as $X_{m}'$ shown in Fig. ~\ref{fig:vifit}.

\noindent \textbf{Decoder.} The decoder is implemented for the vision modality only. Concatenated multimodal representations are fed into the decoder, which is comprised of a projection \textit{Proj($\cdot$)}, GELU activation \textit{GELU($\cdot$)} and Layer Normalization \textit{LayerNorm($\cdot$)}, followed by a linear prediction head \textit{Pred($\cdot$)} of dimension $H_{dim}$:
\begin{equation}
    X'_{fused} = Concat(X'_{c}, X'_{i}, X'_{f})
    \label{eqn:concat}
\end{equation}
\begin{equation}
    \hat{X}_{fused} = Proj(X'_{fused})
    \label{eqn:concat}
\end{equation}
\begin{equation}
    T_{c}^{\prime} = Pred(LayerNorm(GELU(\hat{X}_{fused})))
    \label{eqn:concat}
\end{equation}

\subsection{Training}
\noindent \textbf{Loss Functions.} The task is formulated as a regression problem, which maps camera domain information $T_{c}^{0}$ with first frame only and phone domain data $T_{i}$ and $T_{f}$ to a continuous space in $T_{c}^{\prime}$. Therefore, we employ Mean Squared Error (MSE) as the default loss function by:
\begin{equation}
    L_{MSE} = \frac{1}{N}\sum_{i=1}^{N}(T_{c}^{i}, T_{c}^{\prime i})^{2}
    \label{eqn:mse}
\end{equation}
\noindent where $T_{c}$ is ground truth (GT) and $T_{c}^{\prime}$ is the reconstructed tracklet in a window, $i$ is $BBX$ index in a tracklet and $N$ is the total number of training samples.

Note that our task is to regress bounding boxes by auxiliary information from other modalities in phone domain, due to the cumulative drift error from IMU readings, reconstructed bounding boxes in later frames of a window are more likely to deviate from GT, which results in non-overlapping reconstructed bounding boxes with GT especially at the earlier stage of training. In this case, however, MSE treats each dimension equally without any directional information in terms of where a reconstructed bounding box should move. To alleviate this problem, inspired by Zheng et al. ~\cite{zheng2020distance}, we also investigate Distance-IoU (DIoU) loss function:
\begin{equation}
    L_{DIoU} = \frac{1}{N}\sum_{i=1}^{N} \left(1 - IoU(T^{i}_{c}, T^{\prime i}_{c}) + \frac{\rho^{2}(T^{i}_{c}, T^{\prime i}_{c})}{(s^{i})^{2}} \right)
    \label{eqn:diou}
\end{equation}
\noindent where $\rho$($\cdot$) is the Euclidean Distance between the centroids of $ith$ $BBX$ in $T^{i}_{c}$ and $T^{\prime i}_{c}$ and $s^{i}$ denotes the diagonal length of the smallest enclosing box that covers those two bounding boxes. Training our model by DIoU is more effective in terms of bounding box qualities than MSE since the former guides a model with explicit moving directional information towards the GT, and thus yielded better results in IoU-based metrics shown in Table \ref{tab:loss}~ in later sections. Losses are computed for mini batches during training and therefore $N$ can be the batch size.
Our model is implemented in Python 3.8, PyTorch 1.13.
\footnote{Code is available at \url{https://github.com/bryanbocao/vifit}. Dataset can be downloaded at \url{https://sites.google.com/winlab.rutgers.edu/vi-fidataset/home}.}

\section{Evaluation}
\subsection{Baseline Methods}
We evaluate our approach by comparison against alternative methods. Baseline methods are categorized into two categories: (1) \textbf{traditional handcrafted methods}, including \textit{Broadcasting} (BC), \textit{Pedestrian Dead Reckoning} (PDR) ~\cite{wang2018pedestrian}, and \textit{Kalman Filter} ~\cite{welch1995introduction, li2015kalman}, as well as (2) \textbf{deep learning methods} that includes \baselinemodel~ ~\cite{cao2022vitag}. The details of the baselines are described as follows: \\
\noindent \textbf{\textit{Broadcasting} (BC).} The first frame detections are broadcasted through the rest of missing frames. Note that only the first frame is available, thus linear interpolation cannot be applied without a second frame.
\\
\noindent \textbf{\textit{Pedestrian Dead Reckoning} (PDR)~\cite{li2012reliable, wang2018pedestrian}.} PDR is the process of estimating the current position using previous estimates (the distance traveled and the heading of the pedestrian motion in the North-East-Down (NED) coordinates) obtained from IMU sensors including accelerometer, gyroscope, and magnetometer. 

To estimate the distance, we first determine if the pedestrian is walking or stopping by finding the peaks in the acceleration magnitude from the tri-axial accelerometer data as Boolean values in the vector $sd$, then estimate the distance traveled by assuming that the pedestrian stride length $l$ is 0.6 meters on average. To estimate the phone's heading, we fuse the three sensors data (i.e. accelerometer, gyroscope and magnetometer) to determine the phone’s rotation $\theta$ around the three global axes in NED.
Then the current position $(\hat{x}^t, \hat{y}^t)$ with respect to the previous position $(\hat{x}^{t-1}, \hat{y}^{t-1})$ is computed by
\begin{equation}
          \hat{x}^t = \hat{x}^{t-1}+sd^t \times l \times \cos{\theta^t}
\end{equation}
\begin{equation}
          \hat{y}^t = \hat{y}^{t-1}+sd^t \times l \times \sin{\theta^t}\\
\end{equation}

This results in a history of positions -- the trajectory of the person in the NED coordinates.
To map NED trajectories to image plane, we mark four points on the ground and estimate the meter-to-pixel ratio by Google Maps for outdoor scenes. In the next step, we apply the perspective transformation to map trajectories in NED ($\hat{x}$, $\hat{y}$) to image coordinates ($x'$, $y'$).

Once subjects' feet center positions are obtained, we estimate the other bounding box parameters of width $w$, height $h$ and depth $d$. Inspired by Jiang et al. ~\cite{jiang2021flexible}, we first conduct a data study on the relationship between feet locations and the corresponding bounding boxes' scale (width and height) shown in Fig. ~\ref{fig:datastudy}. Several key observations are made. First, that all scenes share similar relations. Specifically, we do not observe clear pattern between bounding boxes scales and feet center positions in the X-Axis in Fig. ~\ref{fig:datastudy} (a) and (b), while there is a clear linear relationship in the Y-Axis denoted by $*$ in (c) and (d). This is because of the orientation of our camera facing downward to some degrees that subjects closer to the camera appear near the bottom of the camera frame in the vertical axis (Y), which results in larger bounding boxes and vice versa. Second, the linear pattern of $(w, h)$ versus $y$ differs in different scenes, namely the camera orientations vary. The main reason is the difference of camera parameters across scenes, therefore each scene has its specific hyperparameters for this linear relationship.

\begin{figure*}[h]
  \begin{minipage}{1\linewidth}
  \centering
    \subfigure[Widths and pos]{\includegraphics[width=0.161\textwidth]{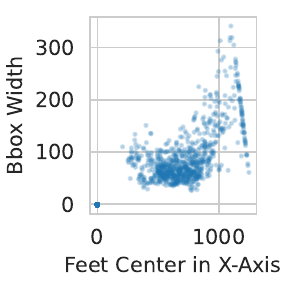}}
    \subfigure[Heights and pos]{\includegraphics[width=0.161\textwidth]{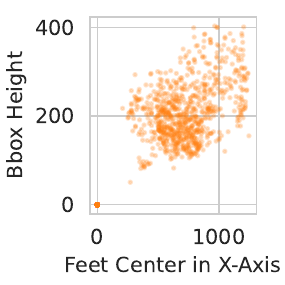}}
    \subfigure[Depths and pos]
    {\includegraphics[width=0.161\textwidth]{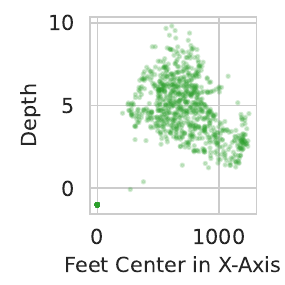}}
    \subfigure[Widths and pos*]
    {\includegraphics[width=0.161\textwidth]{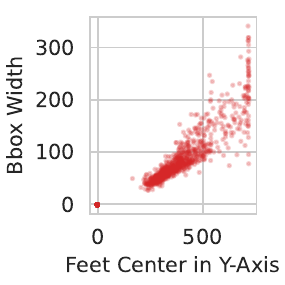}}
    \subfigure[Heights and pos*]
    {\includegraphics[width=0.161\textwidth]{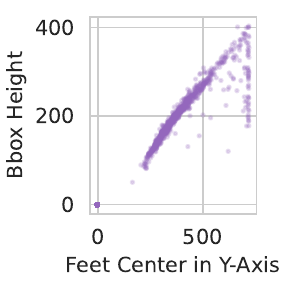}}
    \subfigure[Depths and pos*]
    {\includegraphics[width=0.161\textwidth]{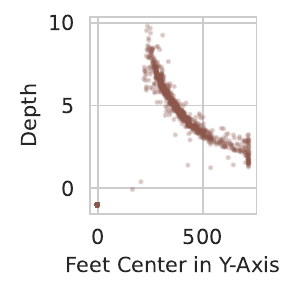}}
  \caption{Relationship between subject coordinates (pos) and bounding box sizes in width and height (pixels) as well as depth (meters) in Scene 4. Observe similar trends across all scenes. * sign indicates the relations described in equation \ref{eqn:w}~-\ref{eqn:d}~.}
  \label{fig:datastudy}
  \end{minipage}
\end{figure*}

Based on the aforementioned observations, we hereby formulate the estimation of bounding boxes' scales ($w'$, $h'$ and $d'$) as a regression problem by functions $f_{1}(y')$, $f_{2}(y')$ and $f_{3}(y')$ of $y$. Formally, $f_{1}$, $f_{2}$ and $f_{3}$ are defined as follows:
\begin{equation}
    w'_{s} = f_{s,1}(y') = k_{s,1} \times y' + b_{s,1}
    \label{eqn:w}
\end{equation}
\begin{equation}
    h'_{s} = f_{s,2}(y') = k_{s,2} \times y' + b_{s,2}
    \label{eqn:h}
\end{equation}
\begin{equation}
    d'_{s} = f_{s,3}(y') = k_{s,3} \times y'^{2} + b_{s,3}
    \label{eqn:d}
\end{equation}
where $k_{s,1}$, $k_{s,2}$, $k_{s,3}$, $b_{s,1}$, $b_{s,2}$, $b_{s,3}$ are the hyperparameters in scene $s \in S$. Finally, we use $x'$, $y'$, $d'$, $w'$ and $h'$ to form the vector of tracklet $T_{c}^{\prime}$ in one frame of a window.

\noindent \textbf{\textit{Kalman Filter} (KF) ~\cite{welch1995introduction, li2015kalman}.}
Kalman filter is widely used in localization and state estimation. Accurate tracking can be attributed to the weight adjustment between measurements and state prediction errors by the importance variable Kalman Gain ($G$). For our tracking task, we adopt a vision-based Kalman filter to estimate a person's bounding box position in an image.

Concretely, a 5-dimensional bounding box vector in the first frame of the current window, as well as a 5-dimensional vector representing velocity in each dimension are formed as the measurement matrix $M$.




The following steps describe one iteration to predict the first frame (frame index = $0$) in the current window $w$ from the previous one $w-1$:

\noindent \textit{(1) State Prediction}:
\begin{equation}\label{eqn:kf_kp}
    \hat{S}^{0,w} = AS^{0,w-1} + Bu^{w} + N^{w}
\end{equation}
where $\hat{S}^{0,w}$ is the prediction of bounding box in the first frame ($0$-indexed). $A$ is the state transition matrix. Note all elements in the diagonal are zeros and the top right 5d matrix is an identity matrix. A prediction considers both bounding box positions and velocities, while the control variable matrix $u^{w}$ is ignored (which is usually acceleration) since there are no control inputs in this task.

\noindent \textit{(2) Covariance Matrix Prediction}:
\begin{equation}\label{eqn:kf_cmp}
    \hat{P}^{0,w} = AP^{0,w-1}A^{T} + Q^{w}
\end{equation}
where $\hat{P}^{0,w}$ is the prediction of covariance matrix in the current window with process noise $Q^{w}$.

\noindent \textit{(3) Kalman Gain}:
\begin{equation}\label{eqn:kf_kg}
    G^{w} = \frac{\hat{P}^{0,w}H^{T}}{H\hat{P}^{0,w}H^{T}+R^{w}}
\end{equation}
Note the general form of Equation \ref{eqn:kf_kg} where $H$ is a matrix to convert $P'$ for $G$'s calculation.

\noindent \textit{(4) State Update}:
\begin{equation}\label{eqn:kf_su}
    S'^{0,w} = \hat{S}^{0,w} + G^{w}[Y - H\hat{S}^{0,w}]
\end{equation}
where $Y=CY_{m} + Z_{m}, Y$ is the measurement matrix converted by matrix $C$ and measurement $Y_{m}$ with measurement noise $Z_{m}$.

\noindent \textit{(5) Covariance Matrix Update}:
\begin{equation}\label{eqn:kf_cmu}
    P^{0,w} = (I - G^{w}H)\hat{P}^{0,w}
\end{equation}


We have experimented initializing all noise matrices by 0-mean Gaussian noise $\mathcal{N}(\mu,\,\sigma^{2})$ where $\mu = 0, \sigma=0.1$, including $N^{w}$, $Q^{w}$, $R^{w}$ and $Z^{w}$. However, results show the empirical IoU per frame in this task is very low (between 0 and 0.1). Therefore we present the Kalman Filter baseline by ignoring all noise terms in our paper. 



Note that in our application, the actual predictions are from only the state in Step \textit{(1) State Prediction}. Assume the system is to reconstruct missing frames in the current window $w$, it first makes a prediction of the state of first frame in the next window $w+1$ by Eq. \ref{eqn:kf_kp} to obtain $\hat{S}^{0,w+1}$, then the final $T_{c}'$ is computed by linearly interpolating $BBX$ between $\hat{S}^{0,w+1}$ and $\hat{S}^{0,w}$.

\noindent \textbf{\baselinemodel~ (X-T) ~\cite{cao2022vitag}.} A multimodal LSTM network in encoder-decoder architecture from ViTag. We follow the training procedure from the publicly released code. Different from the reconstruction path from vision to phone tracklets ($T_c \rightarrow T_{p}'$) in ViTag, we utilize the other reconstruction path from phone data to vision tracklets ($T_{p} \rightarrow T_c'$) to perform the same task in this paper. We also feed only $T_{c}^{0}$ in the first frame into the model. We note the similar related works using recurrent neural network (RNN) or LSTM in the literature, including IONet ~\cite{chen2018ionet, chen2019deep}, milliEgo \cite{lu2020milliego}~, RoNIN ~\cite{herath2020ronin}~, Vi-Fi ~\cite{liu2022vi}~, ViFiCon ~\cite{meegan2022vificon}~, PK-CEN ~\cite{chen2020pedestrian}, FK-CEN ~\cite{xiao2022pedestrian} and so forth, we use \baselinemodel~ ~\cite{cao2022vitag} as the representative of deep learning methods as it is the most closely related to our work.

\subsection{Evaluation Metrics}
Evaluation protocol is consistent across different methods. Specifically, a reconstruct method ($RM$) is a function that takes in (1) $T_{c}^{0} \in \mathbb{R}^{1 \times 5}$ with bounding box detections in the first frame only, (2) $T_{i} \in \mathbb{R}^{WL \times 9}$ and (3) $T_{f} \in \mathbb{R}^{WL \times 2}$ across all frames in a window as input and outputs the reconstructed bounding boxes $T_{c}^{\prime}\in \mathbb{R}^{WL \times 5}$:
\begin{equation}
    T_{c}^{\prime} = RM(T_{c}^{0}, T_{i}, T_{f})
    \label{eqn:rm}
\end{equation}

For each metrics described in later subsections, a reconstruct method ($RM$)'s output $T_{c}^{\prime}$ is compared to the reference of ground truth (GT) $T_{c(GT)}$ in a window.

\noindent \textbf{Minimum Required Frames (MRF).} We introduce MRF as the main metric to measure the smallest number of frames needed for an $MR$ to reconstruct good bounding boxes in a video stream system. The details will be described in the next subsection. \\
\noindent \textbf{Intersection Over Union (IoU).} Since the final output $T_{c}^{\prime}$ consists of bounding boxes, we use Intersection Over Union (IoU) to evaluate how well a reconstructed bounding box matches the GT. IoU is referred to as Jaccard similarity coefficient or the Jaccard index. It is computed by the area of overlap between the reconstructed bounding box and the GT at $ith$ frame divided by their union:
\begin{equation}
    IoU(T_{c}^{i}, T_{c}^{\prime i}) = \frac{|T_{c}^{i} \cap T_{c}^{\prime i}|}{|T_{c}^{i} \cup T_{c}^{\prime i}|}
    \label{eqn:iou}
\end{equation}
\noindent \textbf{Average Precision (AP).} AP is commonly used in object detection in Computer Vision in the standard evaluation tool in COCO ~\cite{lin2014microsoft}. It is calculated by dividing True Positives (TP) by all the positives:
\begin{equation}
    AP@\tau = \frac{TP}{TP+FP}
    \label{eqn:iou}
\end{equation}
where a threshold $\tau$ of IoU is given to determine a TP. It is widely accepted that an IoU score greater than 0.5 is generally considered good. For a thorough evaluation, we also consider the range of IoU thresholds from 0.05 to 0.95 with a step size of 0.05, where AP is computed by first calculating each AP for a specific threshold and then averaging them.

\noindent \textbf{Euclidean Distance (ED).} Euclidean Distance between the predictions from a model and the GT. \\
\noindent \textbf{DIoU Loss.} DIoU loss computed following Equation \ref{eqn:diou}. \\
\noindent \textbf{Depth Correction of FTM (DC$_{f}$).} DC$_{f}$ is used to measure the correction of the depth errors from FTM by $T_{c}^{\prime}$ distance predictions. FTM estimates the distance by Round-trip time (RTT). Analyzed by David ~\cite{houle2021analysis}, the error range of FTM is $1-2$ meters in an ideal case which is the measurement of distance between the phone and the Wi-Fi access point. Since depth is included in the output from our model \nmodel~, it is also interesting to measure to what extent the depth estimation from \model~ is able to correct FTM errors. DF$_{f}$ is computed by
\begin{equation}
    DF_{F} = \epsilon_{Depth} - \epsilon_{FTM}
    \label{eqn:df_f}
\end{equation}
where $\epsilon_{Depth}$ is the error of the depth vector from the prediction $T_{c}^{\prime}$ while $\epsilon_{FTM}$ is the FTM error from depth ground truths.

Unless otherwise specified, each of the aforementioned scores is computed per subject per frame.

\subsubsection{\textbf{Minimum Required Frames}}

Existing common metrics of IoU and AP only focus on the quality of reconstructed bounding boxes in each frame independently. However, it fails to capture frame-related characteristics in video stream systems. In a practical Cam-GPU infrastructure, the number of frames to be dropped to save network bandwidth is of great interest. To capture the upper bound of the number of frames a video can be dropped without compromising the reconstruction performance below a certain threshold, we propose a novel IoU-based metric coined {\em Minimum Required Frames} (MRF). The main intuition of MRF is to capture the minimum number of frames required in order for a method to continuously reconstruct decent bounding boxes in missing frames. Note that MRF is computed in sliding window way while only the first frame in a window is available. Smaller MRF indicates better reconstructions.

Formally, we present the details of MRF in Algorithm \ref{algo:mrf}. Given a video stream $V$ consisting of $F$ frames, window length $WL$ that satisfies $W > 2$ and an IoU threshold $\tau$, the algorithm computes and outputs the {\em Minimum Required Frames} (MRF) for the reconstruct method $RM$. We start by initializing MRF to be $1$ and window stride $(WS)$. Note that two consecutive windows have at least one overlap frame. Then the total number of windows $W=\floor*{\frac{(F-1)}{WS}}$. Then we iterate over all windows with step $WS$. For the first window, we use the ground truth of extracted bounding box in the first frame from the visual modality denoted as $T_{c}^{0}$ (0-indexed), then $RM$ reconstructs bounding boxes in the missing frames in the current window using IMU $T_{i}$ and FTM $T_{f}$ tracklets. When iterating the window at $w$, we first evaluate the quality of reconstructed bounding boxes in the last window by checking if the IoU$^{[w-1]}$ is no less than the predefined threshold $\tau=0.5$. If yes, then we can safely let $RM$ continue reconstructing bounding boxes on top of its previous prediction. More specifically, the last frame in the previous window at $[w-1]$ overlaps with the first frame of the current window at $w$, therefore the reconstructed bounding boxes $T_{c}^{\prime (WL-1)[w-1]}$ (the superscript \textit{WL-1} indicates the last frame of the last window) are used as the input of $T_{c}^{0[w]}$ in the current window $w$ without querying the reference. If IoU is lower than $\tau$, however, we will need to query the reference of bounding boxes and increments MRF by 1. We use annotated bounding boxes as GT given from the dataset to serve as the reference only for evaluation purposes. Note that in practice in real-time streaming, a well-trained visual model such as YOLO can serve as an Oracle model to produce pseudo GT to construct the first frame bounding boxes. The algorithm outputs MRF as the final results. In order to compute IoU score we need a reference of bounding boxes either by GT or pseudo GT from the Oracle model, we will explore confidence-based metrics without reference in future work.

\IncMargin{1em} 
\begin{algorithm}
\caption{Minimum Required Frames \textbf{(MRF)}}
\SetKwData{Left}{left}\SetKwData{This}{this}\SetKwData{Up}{up} \SetKwFunction{Reconstruct}{Reconstruct}
\SetKwInOut{Input}{Input}\SetKwInOut{Output}{Output}
\Input{reconstruct method ${RM}$, video stream $V$, total number of frames $F$, \\
        window length ${WL}$ $s.t.$ $W > 2$, IoU threshold $\tau$ $\leftarrow$ $ 0.5$ \\}
\Output{{\em minimum required frames} MRF}
\tcp*[h]{initialization} \\
$MRF$ $\leftarrow$ $ 1$; \\
$WS$ $\leftarrow$ $ WL - 1$; \tcp*[h]{window stride} \\
$W$ $\leftarrow$ $ \floor*{\frac{(F-1)}{WS}}$; \tcp*[h]{total number of windows} \\

\For{$w \leftarrow 0$ \KwTo $\textbf{W}$}
{
    \If(\tcp*[h]{good reconstructed bbox}){$w > 0$ $\&$ $IoU^{[w-1]}$($T_{c}^{\prime}$, $T_{c}$) $\geq \tau$}{
        \tcp*[h]{continue with last frame at $WL-1$ in the previous window $w-1$} \\
        $T_{c}^{0[w]}$ $\leftarrow$ $ T_{c}^{\prime (WL - 1)[w-1]}$; \\
    }
    \Else{
        $MRF$ $\leftarrow$ $ MRF + 1$; \\
        \tcp*[h]{1st frame bbox GT in the current window $w$} \\
        $T_{c}^{0[w]}$ $\leftarrow$ $ T^{0[w]}_{c(GT)}$; \\
    }
    {$T_{c}^{\prime}$ $\leftarrow$ RM($T_{c}^{0}, T_{i}, T_{f}$); \tcp*[h]{reconstruct bounding boxes in the current window}}
}
\label{algo:mrf}
\end{algorithm}
\DecMargin{1em}

Since the total number of frames in a video stream $F$ varies, MRF will also be changed even if the distribution preserves for the same $RM$. To be invariant to $F$, we hereby introduce {\em Minimum Required Frame Ratio} (MRFR) which is defined as MRF divided by the total number of windows:
\begin{equation}
    MRFR = \frac{MRF}{W}
     \label{eqn:mrf}
\end{equation}

\subsection{Overall Performance}
\begin{figure*}[h]
  \begin{minipage}{1\linewidth}
  \centering
    \subfigure[Minimum Required Frames (MRF)]{\includegraphics[width=0.33\textwidth]{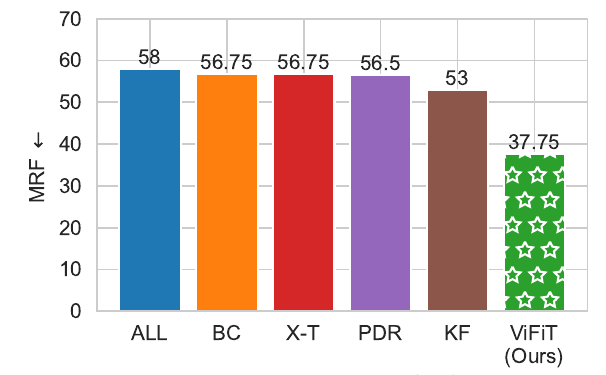}}
    \hspace{42pt}
    \subfigure[Minimum Required Frames Ratio (MRFR)]{\includegraphics[width=0.33\textwidth]{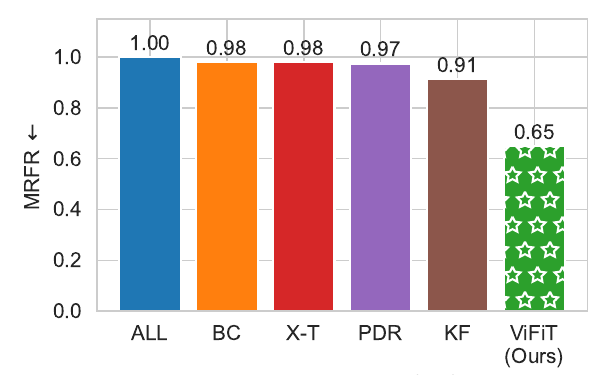}}
  \caption{\textbf{Main} results: comparison of our system \nmodelfff~ (DIoU) against baseline approaches evaluated by {\em Minimum Required Frames} (MRF) shown in (a) and MRFR (Ratio) in (b) for all 4 outdoor scenes with window length $WL=30$ and stride $WS=29$. ALL: the average of the total number of processed windows in one sequence, BC: \textit{broadcasting}, X-T: \baselinemodelfff~, PDR: \textit{Pedestrian Dead Reckoning}, KF: \textit{Kalman Filter}. Observe that most existing methods need to query the first frame. In comparison, \nmodelfff~ reduces frames with only $\frac{37.75}{1683}=2.24\%$ of video frames needed to reconstruct good bounding boxes (IoU = $0.55>0.5$).}
  \label{fig:MRF}
  \end{minipage}
\end{figure*}

Our main result is presented in Fig. \ref{fig:MRF}. In summary, our model \nmodelfff~ (\model~ trained on 30-frame windows) trained by DIoU loss yields the lowest {\em Minimum Required Frames} (MRF) of 37.75 across all 4 outdoor scenes, exceeding the second best method \textit{KF} with 53 by 15.25, demonstrating the effectiveness of our approach in reconstructing bounding boxes for missing frames by fusing phone motion and wireless data. Evaluation is done using window length $WL=30$ and stride $WS=29$. Longer window lengths cover more complicated trajectories such as making turns. Therefore, we evaluate the model \nmodelfff~ with a longer window length than \nmodelf~ in the following study.

To interpret the result of \nmodelfff~ in Fig. \ref{fig:MRF}, $RM$ processes 58 windows ($1 + 58 \times 29 = 1683$ frames) in one scene shown by "ALL", out of which in each of the 37.75 windows \nmodelfff~ queries bounding boxes of the first frame from video on average, resulting in $\frac{37.75}{1683}=2.24\%$ frames used. 
In other words, \nmodelfff~ has generated good bounding boxes for the rest of missing frames, accounting for $1-2.24\%=97.76\%$ of a video. Note the average IoU and AP@.5 of these predictions are $0.55$ and $0.56$ while IoU > 0.5 is generally considered good. This result has demonstrated the effectiveness of our system by exploiting other modalities from phone domain, including IMU and wireless data to save video data.
Compared to IoU or AP, we highlight the practical benefit of MRF that it tells a practitioner a quantitative number to determine the minimum frames (e.g. for saving network transmission) required for reconstruction.  

\begin{figure*}[h]
\begin{center}
\includegraphics[width=0.81\linewidth]{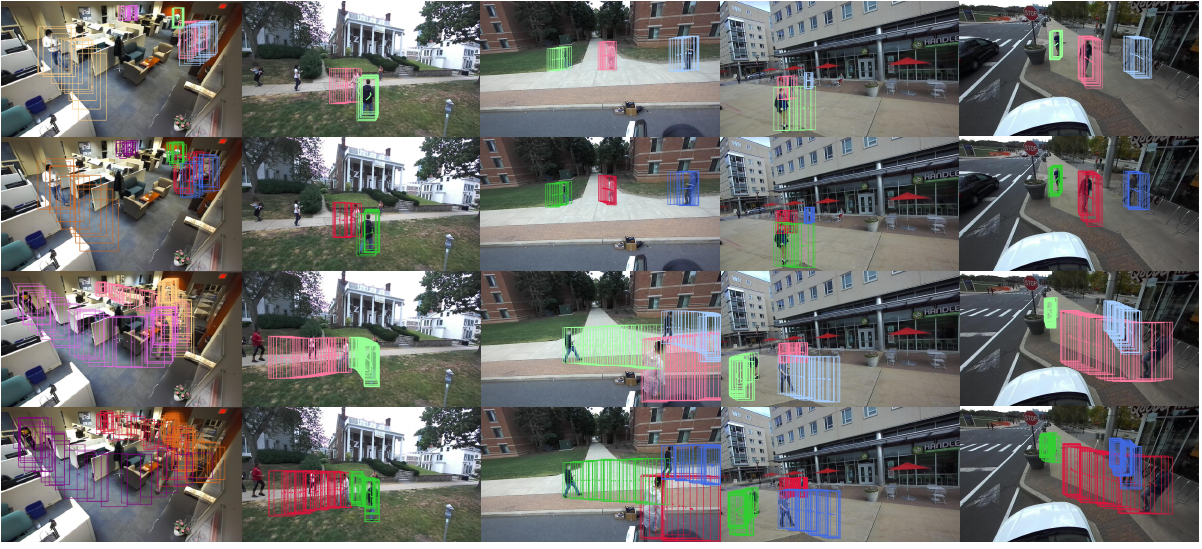}
\end{center}
   \caption{Samples of reconstructed vision tracklets $T_{c}^{\prime}$ and ground truths GT decorated in lighter (1st and 3rd rows) and darker colors (2nd and 4th rows), respectively (Best view in color). Indoor scene is shown in the 1st column while outdoor scenes are displayed from the 2nd to the 5th columns.}
\label{fig:5sc}
\end{figure*}

\noindent \textbf{Continuous Reconstructed Trajectory Length Interval.}
\model~'s performance for continuous reconstructed trajectory length and intervals are shown in Table \ref{tab:interval}. On average, \nmodelfff~ is able to continuously reconstruct decent bounding boxes in 1.55 windows, resulting in 45.01 frames, which correspond to 4.5 seconds with frame rate of 30 FPS. It is worth noting that \nmodelfff~ is capable of reconstructing a continuous trajectory with a maximum length of 87 frames.
Samples are visualized in Fig. \ref{fig:5sc}. For each window, reconstructed bounding boxes are shown in Row 1 and 3. By visual comparison to the ground truth bounding boxes in Row 2 and 4, \nmodel~ is capable of generating decent bounding boxes for missing frames using only a single frame available.

\begin{table}[h]
  \begin{center}
    {\small{
\begin{tabular}{c|cccc}
\toprule
Unit & Avg. & Stdev & Max & Min \\
\midrule
Win & 1.55 & 0.51 & 3 & 1  \\
Frame & 45.01 & 14.77 & 87 & 29  \\
\midrule
Second& 4.5 & 1.48 & 8.7 & 2.9  \\
\bottomrule
\end{tabular}
}}
\end{center}
\caption{Statistics of continuous reconstructed trajectory length and intervals by \nmodelf~ in 4 outdoor scenes with frame rate 10FPS. \nmodelfff~ generate bounding boxes continuously in 45.01 frames (1.55 windows), corresponding to 4.5 seconds with 10FPS frame rate. Note that the maximum continuous trajectory \nmodelfff~ can generate is 87.}
\label{tab:interval}
\end{table}


\subsection{System Analysis}
In this section, we analyze the performance from various system perspectives.

\begin{figure*}[h]
  \begin{minipage}{1\linewidth}
  \centering
    \subfigure[IoU ]{\includegraphics[width=0.32\textwidth]{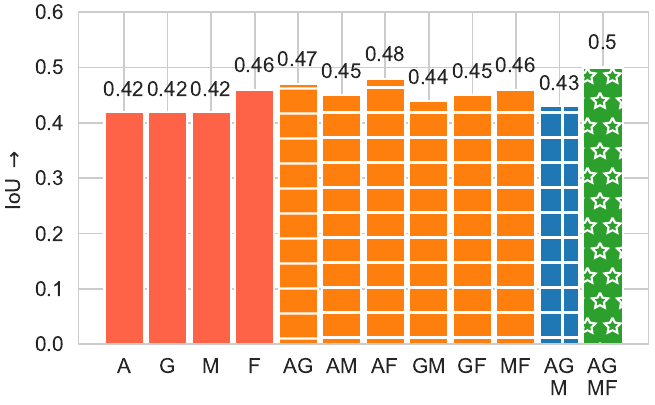}}
    \subfigure[AP@.5]{\includegraphics[width=0.32\textwidth]{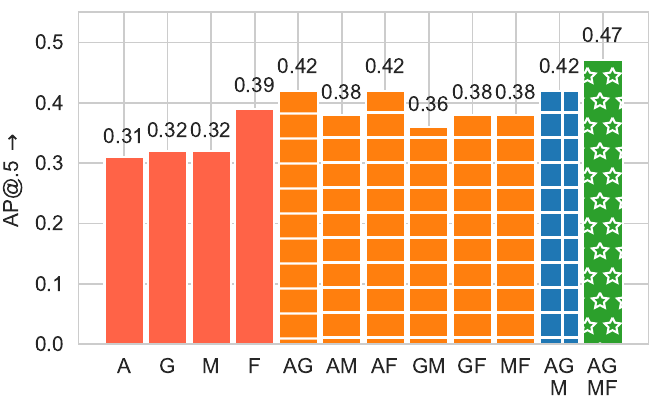}}
    \subfigure[AP@.1]{\includegraphics[width=0.32\textwidth]{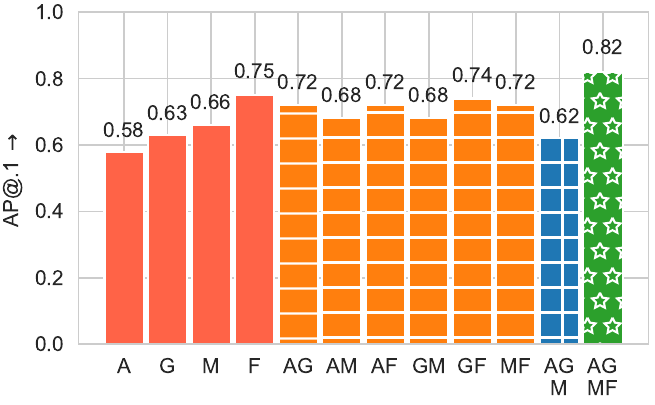}}
  \caption{Phone Feature Ablation Study on \modelfff~ across all 5 scenes. A: Acceleration, G: Gyroscope Angular Velocity, M: Magnetometer Reading, F: FTM. With all phone features (stars in green), \modelfff~ achieves the best performance.}
  \label{fig:ablphone}
  \end{minipage}
\end{figure*}

\noindent \textbf{Phone Feature Ablation Study.} To investigate the effect of phone features, we conduct an ablation study on accelerometer (A), gyroscope (G) and magnetometer (M) readings as well as wireless data FTM (F). Results of all scenes are shown in Fig \ref{fig:ablphone}. Overall, combining all features yields the best performances with IoU, AP@.5 and AP@.1 of 0.5, 0.47 and 0.82, respectively. We observe that a single feature results in poorer performances (light red), while any combination generally helps. F is more useful when combining all A, G and M, demonstrating the design choice of phone features in our system.

\begin{figure}[h]
  \begin{minipage}{1\linewidth}
  \centering
    \subfigure[Effect of H on IoU]{\includegraphics[width=0.23\textwidth]{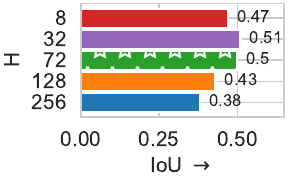}}
    \subfigure[Effect of H on AP@.5]{\includegraphics[width=0.23\textwidth]{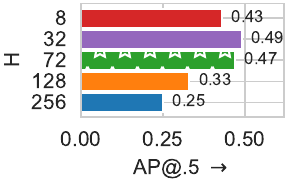}}
    \subfigure[Effect of H on AP@.1]{\includegraphics[width=0.23\textwidth]{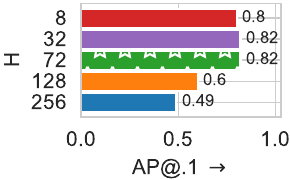}}
    \subfigure[Effect of H on DC$_{f}$]
    {\includegraphics[width=0.23\textwidth]{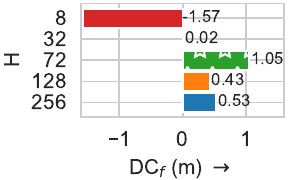}}
  \caption{Effect of $H$ ($\#$ of hidden features). $H=72$ (stars in green) is selected as it achieves the best overall performance considering all 4 metrics. When $H<72$, our model has not enough enough parameters to learn multimodal representations, resulting in lower IoU and being not able to correct DC$_{f}$ (-1.57m). On the other hand, when we increase $H$ to be larger than 72, IoU decreases monotonically, showing a sign of overfitting. Unlike ViT \cite{dosovitskiy2020image}~ that requires pretraining on extremly large datasets, our choice of $72$ demonstrates the effectiveness of the practical training of a transformer on phone data.}
  \label{fig:h}
  \end{minipage}
\end{figure}

\noindent \textbf{Varying Hidden Feature Dimension (H).} To study the effect of hidden feature dimension (H) in the Projection layer, MSA and FeedForward layers shown in the model architecture in Fig \ref{fig:vifit}~, we show the performance of \nmodelfff~ in the following metrics: IoU, AP@.5, AP@.5 and DC$_{f}$. Results in Fig. \ref{fig:h}~ depicts that our choice $H=72$ is a good trade-off, with which \nmodelfff~ achieves the best overall performances across all metrics. It is worth noting that as $H$ increases from 72, 128 to 256, IoU drops monotonically from 0.5, 0.43 down to 0.38, which is a sign of overfitting - a typical problem in training transformers ~\cite{vaswani2017attention} that requires an extremely large number of training samples. However, our study demonstrates a practical training on phone data without extremely large datasets.


\noindent \textbf{Loss Function Study.} To investigate the effect of loss function, we compare our models in window length $WL$ of 10 and 30. Loss functions include {MSE} (Equation \ref{eqn:mse}), {DIoU} (Equation \ref{eqn:diou}), and {DIoU + d} which is the combination of {DIoU} and depth (d) with equal weights. {ED} changes when the definitions of outputs are different, the direct value of {ED} captures the discrepancy (Euclidean Distance) between the reconstructed bounding boxes and GT only in the output space without physical meaning. Therefore we remove {ED} in this study. In each row, a model is trained and tested across all scenes both indoors and outdoors, with window stride $WS=1$. Results show that in general, {MSE} is able to train our models taking all 4 metrics into considerations, while higher IoU is achieved by {DIoU} loss to $0.68$ and $0.52$, compared to $0.65$ and $0.50$ in 10 and 30 window length models. However, training with {DIoU + d} does not bring much benefits to models. One reason can be various ranges between DIoU and depth.
\begin{table}[h]
  \begin{center}
    {\small{
\begin{tabular}{c|c|ccccc}
\toprule
Loss & WL & IoU $\uparrow$ & AP@.5 $\uparrow$ & AP@.1 $\uparrow$ & DC$_{f}$ $\uparrow$ \\
\midrule
MSE & 10F & 0.65 & 0.76 & \textbf{0.96} & \textbf{1.26m} \\
DIoU & 10F & \textbf{0.68} & \textbf{0.78} & \textbf{0.96} & 0.97m \\
DIoU+d & 10F & 0.59 & 0.61 & 0.86 & 1.13m \\
\midrule
MSE & 30F & {0.50} & {0.47} & \textbf{0.82} & \textbf{1.05m}   \\
DIoU & 30F & \textbf{0.52} & \textbf{0.50} & \textbf{0.82} & 0.23m \\
DIoU+d & 30F & 0.47 & 0.43 & 0.72 & 0.49m  \\
\bottomrule
\end{tabular}
}}
\end{center}
\caption{Loss Function Study on \nmodel. Overall, MSE is sufficient to train our model considering IoU-based metrics and depth correction DC$_{f}$ jointly. DIoU can help training a model to reconstruct better bounding boxes with higher IoU score. However, observe that mixing DIoU with depth (d) does not contribute the learning. We hereby use models trained by DIoU for evaluation by MRF.}
\label{tab:loss}
\end{table}

\noindent \textbf{Effect of Window Length ($WL$).} We vary the length of the window, train and test each model by the default MSE loss across all scenes. As $WL$ increases, the shape of trajectories will also be more complicated to reconstruct. Results are obtained by window stride $WS=1$.
\begin{table*}[h]
  \begin{center}
    {\small{
\begin{tabular}{c|cccccccc}
\toprule
Model & IoU $\uparrow$ & AP@.5 $\uparrow$ & AP@.1 $\uparrow$ & $ED$ $\downarrow$ & DIoU Loss $\downarrow$ & DC$_{f}$ $\uparrow$ & $\epsilon_{Depth}$ $\downarrow$ \\
\midrule
\nmodelf~                & \textbf{0.65} &	\textbf{0.76} &	\textbf{0.96} &	\textbf{10.21} & \textbf{0.37} &	\textbf{1.26m} &	\textbf{0.35m} \\
\nmodelfff~         &  0.50 &	0.47 &	0.82 &	11.23 &	0.60 &	1.05m &	0.56m \\ 
\nmodelfffff~          &  0.43 &	0.34 &	0.69 &	11.80 &	0.72 &	0.84m &	0.76m \\
\nmodelfffffff~         & 0.39 &	0.26 &	0.61 &	11.92 &	0.81 &	0.76m &	0.83m \\
\nmodelfffffffff~         & 0.35 &	0.19 &	0.51 &	13.33 &	0.91 &	0.69m &	0.90m \\
\bottomrule
\end{tabular}
}}
\label{tab:main_avg}
\end{center}
\caption{Summary of reconstruction results across all scenes. Each number is averaged across all subjects and frames. Best scores are displayed in bold. DC$_{f}$ is the depth correction from FTM defined as  IoU decreases when $WL$ increases, which is mainly due to more complicated trajectories with longer window lengths.}
\end{table*}



\begin{figure*}[h]
  \begin{minipage}{1\linewidth}
  \centering
    \subfigure[AP for \nmodel~ with different $WL$s]{\includegraphics[width=0.49\textwidth]{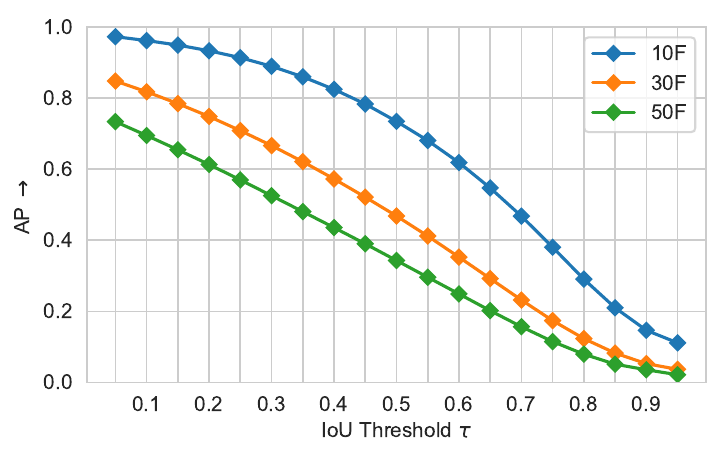}}
    \subfigure[AP for \nmodelfff~ in different scenes]{\includegraphics[width=0.49\textwidth]{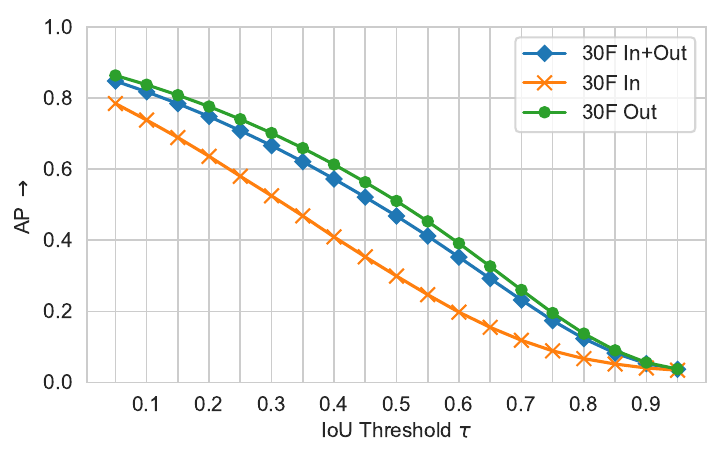}}
   \caption{\nmodelfff~ AP with different IoU thresholds $\tau$ in 0.05:0.95:0.05. Overall, APs are higher with lower thresholds. Longer window length (WL) covers more complicated bounding box trajectories such as making turns, resulting in lower AP (10F to 50F in (a)). Wireless data profile is more complicated in an indoor scene with problems like multipath, yielding lower AP than outdoor scenes (b).}
  \label{fig:iouthreds}
  \end{minipage}
\end{figure*}

\noindent \textbf{AP with Various Thresholds ($\tau$).} AP with different thresholds is displayed in Fig. \ref{fig:iouthreds}. ~\nmodel~ with various window lengths ($WL$) yields different performances. Overall, a higher IoU threshold $\tau$ results in a lower AP since a True Positive (TP) is considered only when its IoU reaches that threshold. Observe a higher performance in smaller $WL$ since it includes less complex trajectories in a shorter time in Fig. \ref{fig:iouthreds} (a). In an open space in outdoor scenes, study by Ibrahim et al. ~\cite{ibrahim2018verification} shows that standard deviation of FTM is positively correlated with distance between the phone and access point, which is the also the distance between a person and the camera. This can benefit a model for better estimation.  As a result, ~\nmodelfff~ achieves higher AP in outdoor scenes than indoors shown in \ref{fig:iouthreds} (b).

\noindent \textbf{Effect of Window Stride ($WS$).} We compare \nmodelf~ and \nmodelf~ with Window Stride $WS=1$ and $WS=WL$ in Fig. \ref{tab:ws}. Observe almost the same results regarding various window strides. We can conclude that in this dataset $WS$ has little impact on the performance of the same model.
\begin{table*}[h]
  \begin{center}
    {\small{
\begin{tabular}{c|cccccccc}
\toprule
Method & IoU $\uparrow$ & AP@.5 $\uparrow$ & AP@.1 $\uparrow$ & ED $\downarrow$ & DIoU Loss $\downarrow$ & DC$_{f}$ $\uparrow$ & $\epsilon_{Depth}$ $\downarrow$ \\
\midrule
\nmodelf~ ($WS=1$)              & {0.65} & 0.76 & {0.96} & {10.21} & 0.37 & 1.26m & {0.35m}  \\
\nmodelf~ ($WS=10$)               & 0.65 & 0.74 & 0.96 & 10.19 & 0.37 & 1.26m & 0.34m  \\ 
\midrule
\nmodelfff~ ($WS=1$)         & {0.50} & {0.47} & {0.82} & {11.23} & {0.60} & {1.05m} & {0.56m}  \\ 
\nmodelfff~ ($WS=30$)         & 0.50 & 0.47 & 0.82 & 11.22 & 0.60 & 1.05m & 0.56m   \\ 
\bottomrule
\end{tabular}
}}
\end{center}
\caption{Sliding window with step $WS=1$ vs $WS=WL$. Results show little impact of $WS$.}
\label{tab:ws}
\end{table*}

\noindent \textbf{Effect of Normalization.} We investigate the effect of normalization on IMU sensors. show in Equation \ref{eqn:norm} \cite{xu2021limu} and Savitzky-Golay smoothing filter ~\cite{savitzky1964smoothing} denoted as (N) and (SG), respectively. For each row, we train and test the model separately. Testing is conducted with window stride $WS=1$ in all scenes. As a result, when increasing the window length ($WL$) from 10F to 90F, we do not observe clear improvements with N or SG, compared to models without normalization or smoothing filtering (w/o NS). We hypothesize that the LayerNorm in \nmodel~ enforces the inputs to be in similar distributions, making it more stable on data with noise, demonstrating the effectiveness of our model on real world data.
\begin{table}[h]
  \begin{center}
    {\small{
\begin{tabular}{p{28pt}|c|ccccc}
\toprule
Method & WL & IoU $\uparrow$ & AP@.5 $\uparrow$ & AP@.1 $\uparrow$ & ED $\downarrow$ & DC$_{f}$ $\uparrow$ \\
\midrule
w/o NS & 10F & 0.42 & 0.30 & 0.62 & 14.63 & 0.83m   \\
SG & 10F & \textbf{0.43}	 & \textbf{0.34}	 & \textbf{0.71}  &	\textbf{11.71}  & 0.79m \\
N &  10F & \textbf{0.43} &	\textbf{0.34} &	0.69 &	11.80 & \textbf{0.84m} \\ 
\midrule
w/o NS & 30F & \textbf{0.51} & \textbf{0.49} & \textbf{0.82} & 11.14 & 1.01m  \\
SG & 30F & 0.48 & 0.44 & 0.79 & 36.33 & 1.00m \\
N &  30F & 0.50 & 0.47 & \textbf{0.82} & 11.23 & \textbf{1.05m} \\ 
\midrule
w/o NS & 50F & \textbf{0.43} & \textbf{0.34} & 0.69 & 11.80 & \textbf{0.84m} \\
SG & 50F & \textbf{0.43} & \textbf{0.34} & \textbf{0.71} & \textbf{11.71} & 0.79m \\
N &  50F & 0.42 & 0.30 & 0.62 & 14.63 & 0.83m \\ 
\midrule
w/o NS & 70F & 0.38 & 0.25 & 0.56 & 14.48 & 0.66m\\
SG & 70F & \textbf{0.39} & \textbf{0.26} & 0.60 & 12.56 & \textbf{0.79m} \\
N &  70F & \textbf{0.39} & \textbf{0.26} & \textbf{0.61} & \textbf{11.92} & 0.76m \\ 
\midrule
w/o NS & 90F & 0.33 & 0.18 & 0.46 & 18.30 & 0.52m \\
SG & 90F & \textbf{0.35} & 0.18 & 0.45 & 19.92 & 0.18m \\
N &  90F & \textbf{0.35} & \textbf{0.19} & \textbf{0.51} & \textbf{13.33} & \textbf{0.69m} \\ 
\bottomrule
\end{tabular}
}}
\end{center}
\caption{Summary of effect of normalization and smoothing on \nmodel~ across all scenes. Each number is averaged across all subjects and frames. Best scores are displayed in bold. Result does not show a clear benefit of normalization or smoothing on IMU sensors. This can be due to the Layer Normalization in \nmodel~ that helps stabilize inputs dynamics from IMU, which proves the practical usage of \nmodel~ on real world data.}
\label{tab:norm}
\end{table}

\subsection{Micro-benchmark}
\begin{figure}[h]
  \begin{minipage}{1\linewidth}
  \centering
    \hspace{-10pt}
    \subfigure[\modelfff~]{\includegraphics[width=0.22\textwidth]{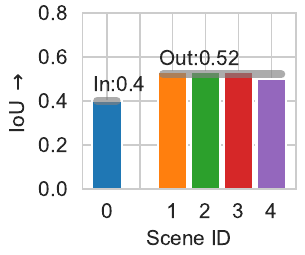}}
    \hspace{55pt}
    \subfigure[\baselinemodelfff~]{\includegraphics[width=0.22\textwidth]{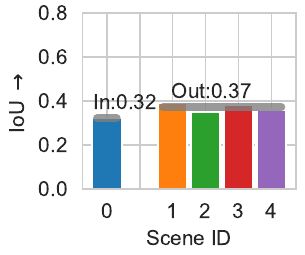}}
  \caption{Scene Study on \nmodelfff~ v.s. \baselinemodelfff~ \cite{cao2022vitag}. Average performances for indoor and outdoor scenes are presented in gray lines. The overall observation is that both deep learning models achieve better performances in outdoor scenes than indoors due to less complex environments. Our model \nmodelfff~ outperforms the state-of-the-art \baselinemodelfff~ in all scenes.}
  \label{fig:scenestudy}
  \end{minipage}
\end{figure}

\noindent \textbf{Scene Analysis.} The data distribution in indoor and outdoor scenes varies. We conduct analysis on each scene separately on $WL=10$ windows with $WS=1$ and compare \nmodelfff~ with the baseline \baselinemodelfff~ \cite{cao2022vitag}. Results are summarized in Fig. \ref{fig:scenestudy}. Overall, both models achieve higher IoU in outdoor scenes (0.52 by \nmodelfff~ and 0.37 by \baselinemodelfff~) than indoors (0.4 and 0.32), respectively. This is mainly due to the more complicated indoor environment than outdoors. In summary, \nmodelfff~ achieves a better performance than \baselinemodelfff~ in all scenes.

\noindent \textbf{Benchmark on Window Length ($WL$).} We conduct benchmarking on various methods by varying window lengths ($WL$) in outdoor scenes. Separate models are trained for different $WL$. Observe that most existing methods fail to produce desirable bounding boxes (IoU > 0.5) when only the first frame is available. Overall, \nmodel~ outperforms all baselines.
\begin{figure}[h]
\begin{center}
   \includegraphics[width=0.55\linewidth]{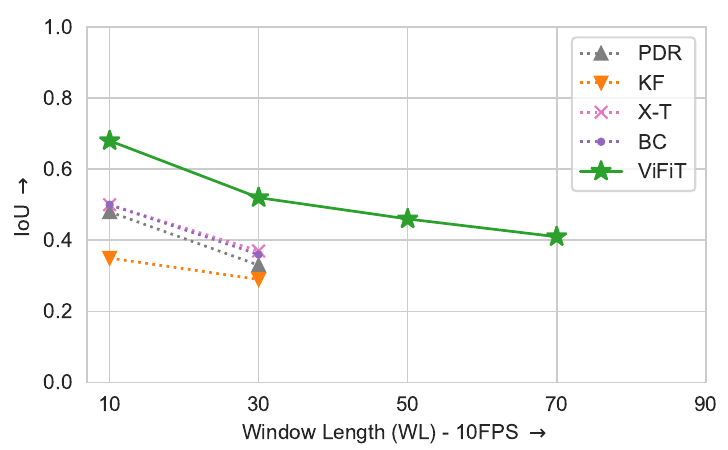}
\end{center}
   \caption{\textbf{Comparison of Methods for Various Window Lengths ($WL$)}. Methods at top right corner are better. Our model \nmodel~ achieves higher AP with longer window lengths compared to other baseline methods. Note that 30 frames correspond to 30 seconds with frame rate 10FPS.}
\label{fig:datasize_dist}
\end{figure}

\noindent \textbf{Analysis on Vision-only Object Detector.} Since our model \nmodel~ can reconstruct bounding boxes with substantially few visual frames from a video, we hereby investigate the extent to which \nmodel~ can preserve high-quality bounding boxes compared to vision-only object detectors. To that end, we evaluate the state-of-the-art vision model's detection when all frames are available, which is the normal inference procedure for a vision model. \nmodel~ is evaluated by the MRF in Algorithm \ref{algo:mrf}~. We emphasize the goal is not to beat vision-only models since they have all the frames available without cumulative drift errors from IMU sensors. Instead, we aim to to analyze two methods in real world settings.

YOLO is selected due to its widely used in various applications. By the time we conduct the experiment, YOLOv5 \cite{glennjocher} is the most recent model.
Results are shown in Table 10. Since IoU is not produced by the YOLOv5 evaluation script, we report AP. Note that AP is positively correlated with IoU since it is based on IoU. Overall, our models \nmodelf~ and \nmodelfff~ preserve decent bounding boxes, achieving an AP@.5 of 0.82 and 0.56 in the outdoor scenes with an extremely small portion of only $2.24\%$ of frames needed in a video, compared to 0.91 and 0.90 by YOLOv5m (medium) and YOLOv5n (nano), respectively. Note that \nmodelf~ and \nmodelfff~ achieve an IoU of 0.71 and 0.55 while IoU>0.5 is generally consider as good. Although our model requires addition IMU and FTM data, it is a good trade-off for the whole system due to the extremely large amount of reduction of video data ($1-2.24\%=97.76\%$ of frames are not used). In addition, our model is lightweight (0.15M \#Params) that does not bring much overhead to existing systems, compared to 1.9M and 21.2M for YOLOv5 nano and medium.

To scrutinize the two systems, one type of error of \nmodel~ mainly comes from the cumulative drift of IMU sensors described in the previous challenges section. IMU sensor readings are in local frame of reference that the relative change of positions and orientations are inferred from accelerations and gyroscope readings. These errors are accumulated in later frames over the trajectory, resulting in more severe bounding box drifts in later frames. In comparison, vision-only model can access the salient part in a frame to precisely locate an object without the cumulative drift.

\begin{table}[h]
  \begin{center}
    {\small{
\begin{tabular}{c|ccccc}
\toprule
Model & AP@.5 $\uparrow$ & Frame Ratio $\downarrow$ & $F$ Reduction $\uparrow$ & Modality & \#Params (M) \\
\midrule
YOLOv5m & 0.91 & 100 \% & 0\% & Vision & 21.2 \\
YOLOv5n & 0.90 & 100 \% & 0\% & Vision & 1.9 \\
\textbf{~\nmodelf} & 0.82 & \textbf{2.25 \%} & \textbf{97.75\%} & Vision, \textbf{IMU, FTM} & \textbf{0.15} \\
\textbf{~\nmodelfff} & 0.56 & \textbf{2.24 \%} & \textbf{97.76\%} & Vision, \textbf{IMU, FTM} & \textbf{0.15} \\
\bottomrule
\end{tabular}
}}
\end{center}
\caption{Analysis with Vision-only model YOLOv5 in outdoor scenes. F: $\#$ Frames, n: nano, m: medium. \nmodelf~ and \nmodelfff~ are able to preserve decent bounding boxes with IoU 0.71 and 0.55 (>0.5 good), respectively. With only $2.24\%$ frames, \nmodelf~ and \nmodelfff~ can maintain AP@.5 as 0.82 and 0.56, demonstrating the good trade-off using phone data (IMU and FTM) to reconstruct bounding boxes with $1-2.24\%=97.76\%$ frame reduction.}
\label{tab:yolo}
\end{table}

\subsection{Real-World Deployment.}
We provide our trained models, script and the docker image upon acceptance. The system can be easily plugged into the common camera to GPU cluster setting by simply appending phone streams to the cluster and run the model. \nmodel~ is lightweight with only $0.15$M parameters without much overhead compared with efficient vision models such as YOLOv5-nano $1.9$M or even ViT Base $86$M.

\section{Conclusion}
In this paper we designed \model~, a system that hosts a transformer based deep learning model to generate tracking information for missing frames in a video. Our work showed that by using strategically selected small number of video frames along with phone's IMU data as well as Wi-Fi FTM can achieve high tracking accuracy. In particular, \model~ uniquely is able to {\em reconstruct} the motion trajectories of human subjects in videos even if the corresponding video frames were not made available or considered lost for processing. To properly evaluate the video characteristics of the system, we propose novel IoU-based metrics {\em Minimum Required Frames} (MRF) and {\em Minimum Required Frames Ratio} (MRFR) for a Camera-GPU system. \model~ achieves $0.65$ MRFR, significantly lower than the second best method \textit{Kalman Filter} of $0.91$ and the state-of-the-art LSTM-based model \baselinemodel~ of $0.98$, resulting in an extremely large amount of frame reduction of $97.76\%$. Through extensive experiments we demonstrated that \model~ is capable of tracking the target with high accuracy and robustness to sensor noises, making it applicable in real world scenarios.


\bibliographystyle{unsrt}  
\bibliography{main}
\clearpage
\newpage

\end{document}